\documentclass[lettersize,journal]{IEEEtran}

\usepackage{easyReview}

\usepackage{amsmath,amsfonts}
\usepackage{array}
\usepackage[caption=false,font=normalsize,labelfont=sf,textfont=sf]{subfig}
\usepackage{textcomp}
\usepackage{stfloats}
\usepackage{url}
\usepackage{verbatim}
\usepackage{graphicx}
\usepackage{cite}
\usepackage{lineno,hyperref}
\usepackage{amssymb}
\usepackage{bm}
\usepackage{mathtools}
\usepackage{color, xcolor}
\usepackage[linesnumbered,ruled,vlined]{algorithm2e}

\usepackage{multirow}

\usepackage{subfig}

\usepackage{overpic}

\usepackage{booktabs}

\usepackage{amssymb}

\usepackage{bbding}

\usepackage{pifont}

\usepackage{threeparttable}
\usepackage{makecell}

\usepackage{threeparttable}
\usepackage{fancyhdr}

\definecolor{myred}{RGB}{255,78,0}    
\definecolor{mygreen}{RGB}{34,139,34} 
\definecolor{myblue}{RGB}{0,0,128}

\begin{document}	
	
	\begin{titlepage}
		\centering
		\vspace*{5cm}
		\textbf{\Large IEEE Copyright Notice}
		\vspace{2cm}
		
		\begin{minipage}{0.8\textwidth}
			\large
			\copyright\ 2025 IEEE. 
			
			Personal use of this material is permitted. Permission from IEEE must be obtained for all other uses, in any current or future media, including reprinting/republishing this material for advertising or promotional purposes, creating new collective works, for resale or redistribution to servers or lists, or reuse of any copyrighted component of this work in other works.
			
			This file corresponds to the accepted version of the manuscript published in IEEE TRANSACTIONS
			ON ROBOTICS, VOL. 41, pp. 5020-5039, 2025. Digital
			Object Identifier 10.1109/TRO.2025.3598145
		\end{minipage}
		\vfill
	\end{titlepage}

\title{AsynEIO: Asynchronous Monocular Event-Inertial Odometry Using Gaussian Process Regression}

\author{Zhixiang Wang, Xudong Li, Yizhai Zhang, Fan Zhang, and Panfeng Huang,~\IEEEmembership{Senior Member,~IEEE}

\thanks{This work was supported by the National Natural Science Foundation of China under Grant 62022067.  (\emph{Corresponding author: Yizhai Zhang.})}


\thanks{The authors are with the Research Center for Intelligent Robotics, School of Astronautics, Northwestern Polytechnical University,  Xi'an 710072, China (e-mail: wangzhixiang@mail.nwpu.edu.cn; lxdli@mail.nwpu.edu.cn; zhangyizhai@nwpu.edu.cn; fzhang@nwpu.edu.cn; pfhuang@nwpu.edu.cn).}
}

\maketitle

\begin{abstract}
Event cameras, when combined with inertial sensors, show significant potential for motion estimation in challenging scenarios, such as high-speed maneuvers and low-light environments. 
While numerous methods exist for producing such estimations, most boil down to solving a synchronous discrete-time fusion problem.
However, the asynchronous nature of event cameras and their unique fusion mechanism with inertial sensors remain underexplored.  
In this paper, we introduce a monocular event-inertial odometry method called AsynEIO, designed to fuse asynchronous event and inertial data within a unified Gaussian Process (GP) regression framework. 
Our approach incorporates an event-driven front-end that tracks feature trajectories directly from raw event streams at a high temporal resolution. 
These tracked feature trajectories, along with various inertial factors, are integrated into the same GP regression framework to enable asynchronous fusion. With deriving analytical residual Jacobians and noise models, our method constructs a factor graph that is iteratively optimized and pruned using a sliding-window optimizer.
Comparative assessments highlight the performance of different inertial fusion strategies, suggesting optimal choices for varying conditions. Experimental results on both public datasets and our own event-inertial sequences indicate that AsynEIO outperforms existing methods, especially in high-speed and low-illumination scenarios.
\end{abstract}


\section{Introduction}
\IEEEPARstart{T}{he} Visual Inertial Odometry (VIO) technology is critical for autonomous robots to navigate in unknown environments. The VIO systems estimate motion states of robots by means of fusing Inertial Measurement Units (IMUs) and visual sensors.
Generally, the IMUs directly measure angular velocities and linear accelerations while the visual sensors observe the bearing information with respect to environments. 
The complementarity of inertial and visual sensors makes the estimation problem more stable in both static and dynamic scenarios. 
Nevertheless, the conventional VIO systems are still prone to crash in several challenging scenarios, including high-speed movements, High-Dynamic-Range (HDR) environments, and sensors with incompatible frequencies.  
Specifically, high-speed movements and HDR environments will cause low-quality grayscale images for intensity cameras and long-period preintegration for IMUs. 
The long-period preintegration rapidly accumulates integration errors, meanwhile, low-quality images intend to deteriorate the visual frondend of VIO systems. Both of them result in a decline in estimation performance. 
In addition, discrete-time estimation and preintegration methods inherently have little support to fuse gyroscopes and accelerometers with incompatible frequencies or even asynchronous measurements. 
Therefore, it is imperative to develop a more robust VIO system to enhance estimation performance in challenging situations.

Event cameras have gained popularity in computer vision and robotics for applications such as object detection \cite{gehrig2024low}, image enhancement \cite{liang2024towards}, and motion estimation \cite{zuo2024cross, Vidal2018Ultimate}.   
As a bio-inspired visual sensor, the event camera has an underlying asynchronous trigger mechanism where the illumination intensity variety of the scenario is independently recorded in a per-pixel manner. 
Benefiting from this special mechanism, event cameras gain prominent performance improvements to conventional cameras, such as low power consumption, low latency, HDR, high-temporal resolution, etc \cite{gallego2020event}. 
Intuitively, it means that event cameras has potential in motion estimation of high-speed and HDR scenarios. 
However, the discrete-time estimation framework adopted in event-based VIO systems greatly hinders the potential of event cameras as illustrated in \cite{wang2024efficient}.

\begin{figure}[!t]
	\centering
	\includegraphics[width=3.2in]{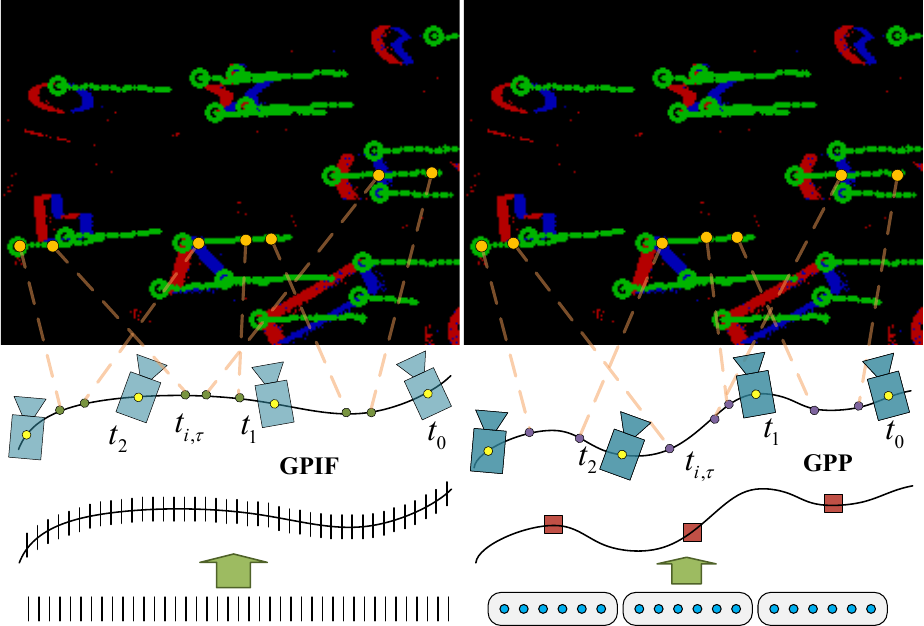}
	\caption{Illustration of two asynchronous fusion schemes (GPIF and GPP). 
		The GPIF conducts a mass of factors for inertial measurements and deforms the continuous-time trajectory in conjunction with GP-based motion prior. The GPP first reduces a series of latent states and then integrates them efficiently with the linear operator at arbitrary timestamps.}
	\label{featuretraj}
\end{figure}

Continuous-time estimation methods have the potential to address foregoing drawbacks of event-based VIO systems.
For these methods, a continuous-time trajectory is firstly defined with splines or kinematic assumptions, and then employed to build the  measurement residuals. 
As the continuous-time trajectory have valid definitions at arbitrary timestamps, it naturally has the capacity of inferring the motion trajectory from totally asynchronous measurements and sensors.
Among them, the Gaussian Process (GP) based methods have gained great interests of researchers for its clearer physical meaning. 
Although previous researchers have proposed to leverage GP-based methods to infer the $SE(3)$ trajectory for event-based VIOs, existing methods still struggle to address the aforementioned challenges. With the introduction of GP-based methods, various inertial fusion schemes combining event cameras and IMUs have emerged. Thus, the interesting question is: which is the best fusion scheme for event-based VIO?

In this paper, we suggest an \textbf{Asyn}chronous \textbf{E}vent-\textbf{I}nertial \textbf{O}dometry (\textbf{AsynEIO}) system to improve the performance in challenging cases and comprehensively assess the reasonable inertial schemes. 
This study significantly extends our previous conference publication \cite{li2024asynchronous} by developing a purely event-based front-end and introducing new inertial fusion schemes.
The proposed AsynEIO utilizes a fully asynchronous fusion pipeline based on GP regression to improve the estimation performance in HDR and high-speed environments.
Two main types of GP-based inertial fusion methods are integrated into the system to permit the fusion of incompatible frequencies or asynchronous inertial measurements (i.e., GP-Inertial Factor (GPIF) and  GP-Preintegration (GPP)), as depicted in Fig.~\ref{featuretraj}.
The GPIF is specifically designed for GP-based continuous-time methods to avoid the necessity of integrating raw measurements. 
We achieve this objective by leveraging a White-Noise-on-Jerk (WNOJ) motion prior and investigating the underlying relationship between the differential $SE(3)$ state variables and raw inertial measurements. 
Instead, the GPP infers data-driven GP models from real-time inertial measurements and generates a series of latent states to replace the motion prior in GPIF.
Inspired by the GPIF, we further extend the conventional discrete Preintegration (Preint) to develop 
a new GP-based inertial factor, termed ExtPreint, where new inertial residuals are directly incorporated into the GP trajectory. 
An additional variant named $\text{GPP}^{\ast}$ is developed to incorporate the original GPP with WNOJ priors.
We fairly compare the performance of these inertial fusion methods under a unified event-inertial pipeline.
Eventually, we conduct experiments on both public datasets and own-collected event-inertial sequences to demonstrate the advantages of the proposal with respect to the state-of-the-art.
In summary, the main contributions of this article are as follows:

\begin{enumerate}{}{}
\item{A fully asynchronous event-inertial odometry, named AsynEIO, 
where a asynchronously parallel, event-triggered front-end is presented to track feature trajectories from raw event streams in real-time.}

\item{Various asynchronous fusion methods are introduced to infer motion trajectories from totally asynchronous measurements in a unified GP-based framework.} 
	
\item{Comprehensive evaluation experiments conducted on public datasets and own-collected sequences demonstrate that the proposed pipeline exhibits superior performance in high-speed scenarios. Informed by the comprehensive evaluation, practical guidance is provided for making reasonable choices of asynchronous fusion schemes under various conditions.}
\end{enumerate}

The remainder of this paper is organized as follows. 
Sec.~\ref{sec:Related Work} reviews the relevant literature. 
The GPIF and GPP are detailed in Sec.~\ref{sec:gpse3} and \ref{sec:gppreintegration}, respectively.  The overall pipeline of AsynEIO is described in Sec.~\ref{sec:eVIO}. Experimental assessments are carried out in Sec.~\ref{sec:experiments}. The discussion of the results is presented in Sec.~\ref{sec:Discussion}. Finally, conclusions are drawn in Sec.~\ref{sec:conclusion}.

\section{Related Work}
\label{sec:Related Work}

Event-based visual-inertial odometry has gained significant research interest for its estimation capabilities in challenging scenarios, such as aggressive maneuvering and HDR environments. A key issue in this field is how to temporally fuse data from
the synchronous, high-rate IMU (e.g., 1 kHz) and the asynchronous
event camera \cite{gallego2020event}. Zhu \emph{et al.} \cite{zihao2017event} aligned batch event packets by two Expectation-Maximization\cite{zhu2017event} steps and fused them with IMU preintegration measurements using an Extended Kalman Filter (EKF). Mahlknecht \emph{et al.} \cite{mahlknecht2022exploring} synchronized measurements from an EKLT tracker \cite{gehrig2020eklt} using extrapolation triggered by a fixed event count. Then, these associated visual measurements were fused with IMU data within a filter-based back-end. However, their front-end schemes incur significant computational overhead.
Rebecq \emph{et al.} \cite{2017Real} proposed to fuse motion compensated event frames and preintegration measurements using a standard keyframe-based framework, where the corner event features were tracked by the vanilla Lucas-Kanade (LK) optical flow. This work was further extended to Ultimate-SLAM \cite{Vidal2018Ultimate} by incorporating standard intensity frames. Lee \emph{et al.} \cite{lee2023event} proposed an 8-DOF warping model to fuse events and frames for accurate feature tracking. The residual terms from the front-end optimization process were used to update an adaptive noise covariance. 
The Time Surface (TS) \cite{lagorce2016hots}, which stores timestamps of recent events, is another viable option for event representation when applying frame-based detection and tracking techniques.
For example, Guan \emph{et al.} \cite{guan2022monocular} adopted Arc* \cite{alzugaray2018asynchronous} for corner detection and implemented LK tracking on the TS. They designed a graph-based back-end to realize tightly-coupled sensor fusion. They further presented PL-EVIO \cite{Guan2023plevio}, which employs multi-modal features (points and Lines) and sensor data (events and frames) to enhance performance in human-made environments.  
Zhou \emph{et al.} \cite{zhou2021event} proposed the first stereo event odometry system  that maximizes spatio-temporal consistency on the TS. 
On this basis, Junkai \emph{et al.} \cite{niu2024imu} designed a gyroscope-based prior to mitigate the degeneracy issue. 
Tang \emph{et al.} \cite{tang2024monocular} suggested an adaptive decay-based TS for feature detecting, where a polarity-aware strategy was applied to enhance robustness, and leveraged the MSCKF state estimator for sensor fusion.
Overall, these methods transform asynchronous event streams into synchronous data associations and convert high-rate IMU data into inter-frame motion constraints through IMU preintegration.

Another category of works processed raw event streams asynchronously  \cite{hu2022ecdt, alzugaray2018asynchronous, alzugaray2020haste, clady2015asynchronous} 
and applied continuous-time state estimation methods to fuse asynchronous data. As demonstrated by Cioffi \emph{et al.} \cite{cioffi2022continuous}, continuous-time estimation methods are normally superior to its discrete counterpart when the sensors are not strictly time-synchronized. Unlike estimating discrete states, continuous-time methods normally assume a continuous trajectory and approximates it through a set of basis functions. By interpolating on continuous trajectories, these methods can relate measurements observed at arbitrary times, such as in multi-camera systems \cite{yang2021asynchronous}, LiDAR odometry \cite{zheng2024traj,dellenbach2022ct,wong2020data} and event cameras \cite{wang2023event,liu2022asynchronous}.
Mueggler \emph{et al.} \cite{mueggler2015continuous} first used cubic B-spline trajectory representation to directly integrate information tracked from line-based maps. They further extended their approach to point-based maps and fused preintegration measurements between adjacent control points to estimate absolute scale \cite{mueggler2018continuous}. 
Lu \emph{et al.} \cite{lu2023event} proposed an event-inertial velometer which leveraged a B-spline continuous-time formulation to fuse the heterogeneous measurements from a stereo event camera and IMU. 
Besides B-spline-based methods, GP regression models the system's motion state using GP and constructs a GP prior between adjacent discrete states, such as  the White-Noise-on-Acceleration (WNOA) prior \cite{barfoot2014batch} and the WNOJ prior \cite{tang2019white}. 
Wang \emph{et al.} \cite{wang2023event} applied motion-compensated RANSAC and a WNOA motion prior to a stereo event odometry pipeline. 
Comparing the two continuous trajectory representations mentioned above,
Johnson \emph{et al.} \cite{johnson2024continuous} demonstrated that the accuracy and computational cost of WNOJ prior GP-based methods and spline-based methods are similar. To the best of our knowledge, there is still a lack of work on GP-based event-inertial odometry using the WNOJ prior.

Recently, researchers have focused on designing new inertial fusion formulas specifically for continuous-time estimation frameworks. 
Several studies have proposed continuous-time formulations for IMU preintegration \cite{le2020gaussian, le2021continuous ,le2023continuous}. These methods use inertial measurements to characterize the system's continuous motion rather than relying on direct priors such as B-splines or GP priors. Le \emph{et al.} \cite{le2020gaussian, le2021continuous} formulated a gyroscope preintegration method, named LPM, where a GP-based latent model was adopted to upsample and infer more precise rotation preintegration. 
This strategy has been employed in event-based VIO systems to fuse IMU data with line features \cite{le2020idol} and corner features \cite{dai2022tightly}. 
IDOL \cite{le2020idol} extracted event clusters of line segments and tracked line features between consecutive event windows. 
They leveraged LPM \cite{le2020gaussian} to associate IMU data with asynchronous event measurements. 
However, this method shown a high computational cost due to its full-batch optimization. Dai \emph{et al.} \cite{dai2022tightly} proposed an asynchronous tracking method relying on decaying event-based corners and adopted a fusion scheme similar to that in \cite{le2020idol}. 
More recently, Gentil \emph{et al.} \cite{le2023continuous} proposed a GP-preintegration (i.e., GPP) method where the latent variable optimization model was applied to analytically infer preintegration measurements and implicitly induce local data-driven GP trajectories. 
They leveraged six independent GP trajectories to address continuous preintegration on $SE(3)$. 
Although their method is specifically designed for GP-based frameworks, it still requires redundant integration and optimization steps rather than directly operating on a GP-based $SE(3)$ trajectory using raw measurements. 
Burnett \emph{et al.} \cite{burnett2024continuous} introduced continuous-time radar-inertial and LiDAR-inertial odometry, adopting a velocity observation model under the WNOA motion prior \cite{anderson2015full} by partially integrating inertial measurements. 
Then, they further improved their method by incorporating angular velocity and body-centric acceleration residuals \cite{burnett2024imu} with the Singer motion prior \cite{wong2020data}. 
However, they reported poor accuracy in their accelerometer measurement model. 
Zheng \emph{et al.} \cite{zheng2024traj} developed a multi-LiDAR and IMU state estimator using two independent GP trajectories on $SO(3)$ and $\mathbb{R}^{3}$. 
In contrast, we present a direct observation model (GPIF and its variant ExtPreint) of raw inertial measurements for the differential states of a unified $SE(3)$ GP trajectory.
In addition, Chamorro \emph{et al.} \cite{chamorro2023event} investigated the event-inertial fusion strategies within a Kalman filter framework, comparing  the conventional ``predict-with-IMU-correct-with-vision'' method against the ``predict-with-model-correct-with-measurement'' approach. 
However, these solutions may face challenges in environments where straight lines are scarce or absent.
Our previous work \cite{li2024asynchronous} investigated two different GP-based methods (i.e., Preint and GPP) to fuse asynchronous visual measurements and inertial preintegration. 
Building on this foundation, we further incorporate more inertial fusion methods within a unified GP-based framework and conduct comprehensive experiments to evaluate their performance in fusing asynchronous measurements. 
Moreover, the previous work relied on intensity images as auxiliary measurements for tracking event streams, limiting its performance in high-speed and HDR scenarios. In AsynEIO, we propose an asynchronous front-end to enable direct processing of pure event streams, achieving excellent adaptability in these challenging conditions. 

\section{GP-Inertial Factor}
\label{sec:gpse3}

The GPIF is directly built upon the continuous-time $SE(3)$ GP trajectory with the WNOJ motion prior. We first define the system kinematics and provide the specific GP regression formulations on the given kinematic states. Then, the observation models relating the differential $SE(3)$ state to raw inertial measurements are presented  along with their analytical Jacobians.

\subsection{Preliminaries and Kinematics}
Define the transformation of the rigid body frame $b$ with respect to (w.r.t) the fixed inertial frame $i$ as a matrix $\boldsymbol{T}_{ib}$, where the pose $\boldsymbol{T}_{ib}$ forms the special Euclidean group, denoted $SE(3)$. Suppose the WNOJ prior to the kinematics of $\boldsymbol{T}_{ib}$ and apply the right multiply update of $SE(3)$. The kinematic differential equation can be expressed as follows:
\begin{align}
	\label{eq:wnoj_prior}
	\dot{\boldsymbol{T}}_{ib} (t) &= \boldsymbol{T}_{ib} (t) (\boldsymbol{\varpi}_{b}^{bi} (t))^\land \nonumber \\
	\dot{\boldsymbol{\varpi}}_{b}^{bi}(t) &= \boldsymbol{s}(t)  \\
	\quad \dot{\boldsymbol{s}}(t) &\sim \boldsymbol{GP}(\boldsymbol{0},\boldsymbol{Q}_c \delta(t-t')) \nonumber,
\end{align}
where 
\begin{equation}
	\label{eq:gen_vel}
	\boldsymbol{\varpi}_{b}^{bi} = \left[\begin{array}{cc}
		\boldsymbol{\omega}_{b}^{bi} \\ \boldsymbol{\nu}_{b}^{bi} 
	\end{array}
	\right]
\end{equation}
is the generalized velocity, $\boldsymbol{s}$ is the generalized acceleration, and $\boldsymbol{Q}_{c}$ is the power spectral
density matrix of the GP. The above equations are consistent with that in \cite{tang2019white}, except that we adopt a right-perturbation update convention.
Given the detailed rotation and translation components as
\begin{equation}
	\label{eq:tw_def}
	\boldsymbol{T}_{ib} =
	\begin{bmatrix}
		\boldsymbol{C}_{ib} & \boldsymbol{r}_{i}^{bi}\\
		\boldsymbol{0}^{\top} & 1
	\end{bmatrix}, \quad 
	{\boldsymbol{\varpi}_{b}^{bi}}^{\land} =
	\begin{bmatrix}
		(\boldsymbol{\omega}^{bi}_{b})^{\land} & \boldsymbol{\nu}_{b}^{bi}\\
		\boldsymbol{0}^{\top} & 0
	\end{bmatrix},
\end{equation}
where $(\boldsymbol{\omega}^{bi}_{b})^{\land}$ maps the angular velocity vector to a skew-symmetric matrix. Then, we have $\dot{\boldsymbol{r}}_{i}^{bi} = \boldsymbol{C}_{ib} \boldsymbol{\nu}_{b}^{bi}$. Comparing the subscripts, $\dot{\boldsymbol{r}}_{i}^{bi}$ can be defined as $ \boldsymbol{\nu}_{i}^{bi}$, which represents the linear velocity of body frame w.r.t the inertial frame observed and expressed in the inertial frame. 
In fact, the generalized linear velocity $\boldsymbol{\nu}_{b}^{bi}$ indicates the same linear velocity observed in the inertial frame but expressed in the body frame. 
Notice that $\boldsymbol{\nu}_{b}^{bi} \neq \dot{\boldsymbol{r}}_{b}^{bi}$, where $\dot{\boldsymbol{r}}_{b}^{bi}$ represents the linear velocity of body frame w.r.t the inertial frame observed and expressed in the body frame.
Thereby, the generalized acceleration is given by
\begin{equation}
	\label{eq:gen_acc}
	\dot{\boldsymbol{\varpi}}_{b}^{bi} = \left[\begin{array}{cc}
		\dot{\boldsymbol{\omega}}_{b}^{bi} \\
		\dot{\boldsymbol{\nu}}_{b}^{bi}
	\end{array}\right]  
	= \left[\begin{array}{cc}
		\dot{\boldsymbol{\omega}}_{b}^{bi} \\
		\dot{\boldsymbol{C}}_{bi} \dot{\boldsymbol{r}}_{i}^{bi} + \boldsymbol{C}_{bi} \ddot{\boldsymbol{r}}_{i}^{bi}
	\end{array}\right],
\end{equation}
where 
$\dot{\boldsymbol{C}}_{bi} = \boldsymbol{C}_{bi} (\boldsymbol{\omega}_{i}^{ib})^{\land}$. We know that $(\boldsymbol{C} \boldsymbol{\omega})^{\land} \equiv \boldsymbol{C} (\boldsymbol{\omega})^{\land} \boldsymbol{C}^{\top}$ and $\boldsymbol{\omega}_{b}^{bi} = - \boldsymbol{C}_{bi} \boldsymbol{\omega}_{i}^{ib} $. Substituting them into \eqref{eq:gen_acc}, we can show  
\begin{equation}
\label{eq:v_dot}
\dot{\boldsymbol{\nu}}_{b}^{bi} = - (\boldsymbol{\omega}^{bi}_{b})^{\land} \boldsymbol{\nu}_{b}^{bi} + \boldsymbol{C}_{bi} \ddot{\boldsymbol{r}}_{i}^{bi},
\end{equation}
where $\ddot{\boldsymbol{r}}_{i}^{bi}$ represents the linear acceleration of body frame w.r.t the inertial frame observed and expressed in the inertial frame, and $\boldsymbol{\omega}_{b}^{bi}$ is the angular velocity of body frame w.r.t the inertial frame observed in the body frame. The kinematic state can be defined as 
\begin{equation}
	\label{eq:kin_state}
	\boldsymbol{x}(t) = \{\boldsymbol{T}(t), \boldsymbol{\varpi}(t), \dot{\boldsymbol{\varpi}}(t)\},
\end{equation}
where the sub- and superscripts have been dropped and \eqref{eq:wnoj_prior}-\eqref{eq:v_dot} describe the Stochastic Differential Equations (SDEs) of kinematics. Since $\boldsymbol{x}(t) \in SE(3) \times \mathbb{R}^{12}$ forms a manifold, these nonlinear SDEs are difficult to directly deploy in continuous-time estimation. 

\subsection{GP Regression on Local Linear Space}
The following derivations follow \cite{wong2020data}, except that we employ a right-perturbation.
To linearize the nonlinear SDEs, the local variable $\boldsymbol{\xi}$ in the tangent space of $\boldsymbol{T}_{k}$ can be given by
\begin{equation}
	\label{eq:Tlocal}
	\boldsymbol{T}(t) = \boldsymbol{T}_k \exp((\boldsymbol{\xi}_k(t))^\land),
\end{equation}
where $\boldsymbol{T}_{k} \doteq \boldsymbol{T}(t_{k})$, and $\exp(\bullet)$ is the exponential map from $\mathfrak{se}(3)$ to $SE(3)$. The time derivative of \eqref{eq:Tlocal} indicates the relationship between local variable and kinematic state as 
\begin{equation}
	\label{eq:rel_local_velocity}
	\boldsymbol{\varpi}(t) = \boldsymbol{\mathcal{J}} (\boldsymbol{\xi}_k(t))\dot{\boldsymbol{\xi}}_k(t),
\end{equation}
where $\boldsymbol{\mathcal{J}} (\boldsymbol{\xi}_k(t))$ is the right Jacobian of $\boldsymbol{\xi}_k(t)$. Notice that $\boldsymbol{\mathcal{J}} \approx \boldsymbol{1}^{6\times 6}$ when $\boldsymbol{\xi}_k(t)$ is small enough.  Further take the time derivative of \eqref{eq:rel_local_velocity} and apply the first-order approximation, we can show the relationship between $\boldsymbol{\varpi}(t)$ and $\ddot{\boldsymbol{\xi}}_{k}(t)$: 
\begin{equation}
	\label{eq:ddxi}
	\ddot{\boldsymbol{\xi}}_{k}(t) \approx \boldsymbol{\mathcal{J}}(\boldsymbol{\xi}_{k}(t))^{-1}
	\dot{\boldsymbol{\varpi}}(t) + \frac{1}{2}(\boldsymbol{\mathcal{J}}(\boldsymbol{\xi}_{k}(t))^{-1} \boldsymbol{\varpi}(t) )^{\curlywedge} \boldsymbol{\varpi}(t),
\end{equation}
where $(\bullet)^{\curlywedge}$ is the adjoint matrix operator. For our definition, we have
\begin{equation}
	\boldsymbol{\xi}^{\curlywedge} = \left[
	\begin{array}{cc}
		\boldsymbol{\phi}\\
		\boldsymbol{\rho}
	\end{array}
	\right]^{\curlywedge} = \left[
	\begin{array}{cc}
		\boldsymbol{\phi}^{\land} & \boldsymbol{0} \\
		\boldsymbol{\rho}^{\land} & \boldsymbol{\phi}^{\land}
	\end{array}
	\right],
\end{equation} 
where $\boldsymbol{\phi} \in \mathbb{R}^{3}$ is the rotation component and $\boldsymbol{\rho} \in \mathbb{R}^{3}$ is the translation component. Now, the local linear kinematic state can be defined  w.r.t $\boldsymbol{T}_{k}$ as
\begin{equation}
	\label{eq:localstate}
	\boldsymbol{\gamma}_k(t) \triangleq \left[
	\begin{array}{ccc}
		\boldsymbol{\xi}_k(t)\\
		\dot{\boldsymbol{\xi}}_k(t) \\
		\ddot{\boldsymbol{\xi}}_{k}(t)
	\end{array}
	\right].
\end{equation}
Similarly, we can assume the motion prior to be  
\begin{equation}
	\dddot{\boldsymbol{\xi}}_{k}(t) \sim \boldsymbol{GP}(\boldsymbol{0},\boldsymbol{Q}_c \delta(t-t')).
\end{equation}
On this basis, the linear time-invariant SDEs are
\begin{equation}
	\dot{\boldsymbol{\gamma}}_{k}(t) = 
	\underbrace{\begin{bmatrix} 
		\boldsymbol{0} & \boldsymbol{1} & \boldsymbol{0} \\
		\boldsymbol{0} & \boldsymbol{0} & \boldsymbol{1} \\
		\boldsymbol{0} & \boldsymbol{0} & \boldsymbol{0}
	\end{bmatrix}}_{\boldsymbol{A}} \boldsymbol{\gamma}_{k}(t) + 
	\underbrace{\begin{bmatrix}
		\boldsymbol{0} \\
		\boldsymbol{0} \\
		\boldsymbol{1}
	\end{bmatrix}}_{\boldsymbol{L}} \dddot{\boldsymbol{\xi}}_{k}(t),
\end{equation}
which has a closed-form solution for the state transition function $\boldsymbol{\Phi}(t, t_{k}) = \exp(\boldsymbol{A}(t-t_{k}))$. Meanwhile, the transition equation of the covariance matrix  can be given by
\begin{equation}
	\label{eq:covar_mat}
	\boldsymbol{Q}_{k}(t) = \int_{t_{k}}^{t} \boldsymbol{\Phi}(s, t_{k}) \boldsymbol{L} \boldsymbol{Q}_{c} \boldsymbol{L}^{\top} \boldsymbol{\Phi}(s, t_{k})^{\top} ds.
\end{equation}

\subsection{WNOJ Prior Residual}
Under the WNOJ assumption, the GP prior residual between timestamp $t_k$ and $t_{k+1}$ can be defined as
\begin{equation}
	\label{PriorRes}
	\boldsymbol{e}_{k} = \left[\begin{array}{ccc}
		\Delta t_k\boldsymbol{\varpi}_k + \frac{1}{2} \Delta t_{k}^{2} \dot{\boldsymbol{\varpi}}_{k}  - \boldsymbol{\xi}_{k,k+1}\\
		\boldsymbol{\varpi}_k + \Delta t_{k} \dot{\boldsymbol{\varpi}}_{k} -\boldsymbol{\mathcal{J}}_{k,k+1}^{-1}\boldsymbol{\varpi}_{k+1} \\
		\dot{\boldsymbol{\varpi}}_{k}-\boldsymbol{\mathcal{J}}_{k,k+1}^{-1} \dot{\boldsymbol{\varpi}}_{k+1}  -\frac{1}{2}(\boldsymbol{\mathcal{J}}_{k,k+1}^{-1} \boldsymbol{\varpi}_{k+1} )^{\curlywedge} \boldsymbol{\varpi}_{k+1}
	\end{array}
	\right],
\end{equation}
in which $\Delta t_{k} = t_{k+1} - k_{k}$,  $\boldsymbol{\xi}_{k,k+1} = \log(\boldsymbol{T}_k^{-1}\boldsymbol{T}_{k+1})^\vee$, $\boldsymbol{\mathcal{J}}_{k,k+1} = \boldsymbol{\mathcal{J}}(\boldsymbol{\xi}_{k,k+1})$, and $\boldsymbol{\varpi}_{k} = \boldsymbol{\varpi}(t_{k})$. Notice that $\log(\bullet)$ and $(\bullet)^{\vee}$ are the inverse maps of $\exp(\bullet)$ and $(\bullet)^{\land}$, respectively. We can compute the covariance matrix for prior residual from \eqref{eq:covar_mat}. Furthermore, the detailed Jacobians of prior residuals are listed in Appendix-B.

\subsection{GP-Inertial Residual and Jacobian}

Generally, the measurement of  the gyroscope is the noisy angular velocity $\tilde{\boldsymbol{\omega}}_{b}^{bi} =\boldsymbol{\omega}_{b}^{bi} + \mathbf{b}_{g} + \boldsymbol{\varsigma}_{g}$, and  the measurement of the accelerometer $\tilde{\boldsymbol{a}}$ is the noisy value 
\begin{equation}
	\label{eq:acc_measure}
	\tilde{\boldsymbol{a}} = \boldsymbol{C}_{bi} (\ddot{\boldsymbol{r}}_{i}^{bi} -\boldsymbol{g}_{i} ) + \boldsymbol{b}_{a} + \boldsymbol{\varsigma}_{a},
\end{equation}
where $\boldsymbol{\varsigma}_{g} \sim \mathcal{N}(\boldsymbol{0},\boldsymbol{Q}_{g})$ and  $\boldsymbol{\varsigma}_{a}  \sim \mathcal{N}(\boldsymbol{0},\boldsymbol{Q}_{a})$ are the Gaussian measurement noises, $\boldsymbol{g}_{i}$ is the gravitational acceleration,  $\mathbf{b}_{g}$ and $\mathbf{b}_{a}$ are biases that obey the Brownian movement.
\begin{align}
	\dot{\boldsymbol{b}}_g \sim \mathcal{N}(\boldsymbol{0}, \boldsymbol{Q}_{bg}), \\
	\dot{\boldsymbol{b}}_a \sim \mathcal{N}(\boldsymbol{0}, \boldsymbol{Q}_{ba}),
\end{align}
where $\boldsymbol{Q}_{a}, \boldsymbol{Q}_{g}, \boldsymbol{Q}_{bg}, \boldsymbol{Q}_{ba}$ are covariance matrices. 

Before applying the GPIF, we need first know the kinematic state at an arbitrary timestamp.  To query the trajectory at timestamp $\tau$ between $t_{k}$ and $t_{k+1}$, we firstly interpolate the local variable $\boldsymbol{\gamma}_k(\tau)$ \cite{wong2020data} with
\begin{align}
	\label{eq:query_lin_state}
	\boldsymbol{\gamma}_k(\tau)&=\boldsymbol{\Lambda}(\tau)\boldsymbol{\gamma}_k(t_k) + \boldsymbol{\Psi}(\tau)\boldsymbol{\gamma}_k(t_{k+1}), \nonumber \\
	\boldsymbol{\Lambda}(\tau)&=\boldsymbol{\Phi}(\tau,t_k)-\boldsymbol{\Psi}(\tau)\boldsymbol{\Phi}(t_{k+1},t_k), \nonumber \\
	\boldsymbol{\Psi}(\tau)&=\boldsymbol{Q}_{k}(\tau) \boldsymbol{\Phi}(t_{k+1},\tau)^{\top}\boldsymbol{Q}_{k}(t_{k+1})^{-1}.
\end{align}
Then, in conjunction with \eqref{eq:Tlocal}-\eqref{eq:localstate}, the kinematic state $\boldsymbol{x}(\tau)$ in \eqref{eq:kin_state} can be resolved.

With the GP-based continuous-time trajectory, we construct the GPIF  directly from the angular velocity and linear acceleration states, which avoids the preintegration. Considering \eqref{eq:gen_vel}, \eqref{eq:v_dot} and \eqref{eq:acc_measure}, the residual of GPIF can be given by 
\begin{align}
	\label{eq:inertial_residual}
\boldsymbol{e}_{g}(\tau) &= 
	\tilde{\boldsymbol{\omega}}(\tau) - \boldsymbol{\omega}(\tau) - \boldsymbol{b}_{g}(\tau),	\nonumber\\
\boldsymbol{e}_{a}(\tau) &=	\tilde{\boldsymbol{a}}(\tau) -\dot{\boldsymbol{\nu}}(\tau) - (\boldsymbol{\omega}(\tau))^{\land} \boldsymbol{\nu}(\tau) +  \boldsymbol{C}(\tau)^{\top} \boldsymbol{g} - \boldsymbol{b}_{a}(\tau),
\end{align}
where the sub- and superscripts have been dropped. Neglecting the effects of bias and gravity, we can see that the raw acceleration measurement $\tilde{\boldsymbol{a}}$ has a direct relationship with the kinematic state $\dot{\boldsymbol{\nu}}$ and an extra term $\boldsymbol{\omega} ^{\land} \boldsymbol{\nu}$. Notably, when the angular velocity or linear velocity is very small, or when the axis of rotation is approximately aligned with the direction of the linear velocity, this extra term can be neglected. In other cases, this extra term should be taken into account. Up to this point, we have established a detailed residual model between the raw IMU measurements and the kinematic states.

The covariance matrix of the direct residual is exactly the measurement covariance $diag (\boldsymbol{Q}_{g}, \boldsymbol{Q}_{a} )$. The Jacobians of GPIF residuals can be derived from the chain rule as follows:
\begin{equation}
	\frac{\partial \boldsymbol{e}(\tau)}{\partial \boldsymbol{x}(t_k)} = \frac{\partial \boldsymbol{e}(\tau)}{\partial \boldsymbol{x}(\tau)} \frac{\partial \boldsymbol{x}(\tau)}{\partial \boldsymbol{x}(t_{k})},
\end{equation}
where the derivations of $\frac{\partial \boldsymbol{x}(\tau)}{\partial \boldsymbol{x}(t_{k})}$ can be found in the Appendix-C. Here, the nonzero components of $\frac{\partial \boldsymbol{e}(\tau)}{\partial \boldsymbol{x}(\tau)}$ are listed as follows:
\begin{align}
\frac{\partial \boldsymbol{e}_{g}(\tau)}{\partial \boldsymbol{\omega}(\tau)} &=
\frac{\partial \boldsymbol{e}_{g}(\tau)}{\partial \boldsymbol{b}_{g}(\tau)} = 
\frac{\partial \boldsymbol{e}_{a}(\tau)}{\partial \dot{\boldsymbol{\nu}}(\tau)} = \frac{\partial \boldsymbol{e}_{a}(\tau)}{\partial \boldsymbol{b}_{a}(\tau)} = -\boldsymbol{1}, \nonumber\\
\frac{\partial \boldsymbol{e}_{a}(\tau)}{\partial \boldsymbol{\nu}(\tau)} &= -(\boldsymbol{\omega}(\tau))^{\land}, \ 
\frac{\partial \boldsymbol{e}_{a}(\tau)}{\partial \boldsymbol{\omega}(\tau)} = (\boldsymbol{\nu}(\tau))^{\land}, \nonumber\\ 
\frac{\partial \boldsymbol{e}_{a}(\tau)}{\partial \boldsymbol{C}(\tau)} &= \left(\boldsymbol{C}(\tau)^{\top} \boldsymbol{g}\right)^{\land}, 
\end{align}
where $\boldsymbol{b}_{a}(\tau)$ and $\boldsymbol{b}_{g}(\tau)$ are calculated with linear interpolation.
In addition, a bias prior factor is attached between each pair of sampling timestamps $t_{k}$ and $t_{k+1}$, as follows:
\begin{align}
	\label{eq:bias_prior}
	\boldsymbol{e}_{ba}(t_{k}) = \boldsymbol{b}_{a}(t_{k+1}) - \boldsymbol{b}_{a}(t_{k}), \nonumber\\
	\boldsymbol{e}_{bg}(t_{k}) = \boldsymbol{b}_{g}(t_{k+1}) - \boldsymbol{b}_{g}(t_{k}).
\end{align}

\section{GP-Preintegration} \label{sec:gppreintegration}
The GPP provides a different insight to utilize GP regression framework to tightly fuse inertial and other asynchronous measurements. Rather than conducting the kinematic prior, the GPP assumes local inertial measurements as a group of independent GPs and instantiates them on-the-fly in a data-driven manner \cite{le2023continuous}. 

\subsection{GP Assumptions on Local Measurements}
Concretely, the GPP models the ``local accelerations'' $\boldsymbol{a}_{k}(t) \in \mathbb{R}^{3}$ and the ``local angular velocities'' $\dot{\boldsymbol{\phi}}_{k}(t) \in \mathbb{R}^{3}$ as six independent GPs. Therefore, we have
\begin{equation}
	\label{eq:UGPM GP}
	\begin{aligned}
		\dot{\boldsymbol{\phi}}_{k}(t)  \sim \boldsymbol{GP}(0,k_{\phi}(t,t_{k}) \boldsymbol{I}_{3} ),
		\\
		\boldsymbol{a}_{k}(t)  \sim \boldsymbol{GP}(0, k_a(t,t_{k}) \boldsymbol{I}_{3} ),
	\end{aligned}
\end{equation}
where $\dot{\boldsymbol{\phi}}_{k}(t) \in \mathbb{R}^{3}$ is the time derivative of $\boldsymbol{\phi}_{k}(t)$, and $k_{\phi}(t,t_{k}),k_{a}(t,t_{k}) $ are kernel functions of GPs, and $\boldsymbol{I}_{3} \in \mathbb{R}^{3\times 3}$ is an identity matrix. The local variables are defined as
\begin{equation}
	\label{eq:localgppval}
	\begin{aligned}
		\boldsymbol{\phi}_{k}(t) &= \log( \boldsymbol{C}(t_{k}) ^{\top} \boldsymbol{C}(t)  ) ^{\vee}, \\
		\boldsymbol{a}_{k}(t) &= \exp(\boldsymbol{\phi}_{k}(t)^{\land}) \boldsymbol{a}(t),
	\end{aligned}
\end{equation}
in which $\boldsymbol{C}(t)$ is the rotation matrix of the body frame with respect to the inertial frame at the timestamp $t$, and $\boldsymbol{a}(t)$ is the so-called proper acceleration of the body frame. From the definition of \eqref{eq:acc_measure}, we know that  $\boldsymbol{a}(t) = \boldsymbol{C}_{bi} (\ddot{\boldsymbol{r}}_{i}^{bi} -\boldsymbol{g}_{i} )$. As shown in \eqref{eq:UGPM GP} and  \eqref{eq:localgppval}, the local accelerations and angular velocities are expressed in a local body frame at the timestamp $t_{k}$. On this basis, these variables can be integrated linearly.

\subsection{GP-Preintegration on Induced Latent States} \label{sec:gpp_theory}
To permit efficient analytical preintegration, the GPP introduces a series of the latent states as the induced pseudo-observations and conducts an online optimization problem to infer them from inertial measurements. With the linear operator \cite{sarkka2011linear} defined on corresponding latent states, it can effectively execute the preintegration for given timestamps.

Suppose the latent states $\hat{\boldsymbol{\rho}}_{j}$ and $\hat{\boldsymbol{\alpha}}_{j}$  be noisy observed vectors of local variables at the timestamp $t_j$ as in \eqref{eq:UGPM GP}, we have
\begin{equation}
	\label{UGPM meas}
	\begin{aligned}
		\tilde{\boldsymbol{\rho}}_j &= \dot{\boldsymbol{\phi}}_{k}(t_{j}) + \boldsymbol{\eta}_{\rho} \ with \  \boldsymbol{\eta}_{\rho} \sim \mathcal{N}(0, \sigma_{\phi}^2 \boldsymbol{I}_{3}),
		\\
		\tilde{\boldsymbol{\alpha}}_j &= \boldsymbol{a}_{k}(t_{j}) + \boldsymbol{\eta}_{\alpha} \ with\  \boldsymbol{\eta}_{\alpha} \sim \mathcal{N}(0,\sigma_{a}^2 \boldsymbol{I}_{3}),
	\end{aligned}
\end{equation}
where $\boldsymbol{\eta}$ follows the zero-mean Gaussian distribution with a diagonal covariance matrix. 
By conditioning upon these latent states, the posterior values of local variables in \eqref{eq:UGPM GP} can be expressed as 
\begin{equation}
	\label{eq:posteriorGP}
	\begin{aligned}
		\tilde{\dot{\boldsymbol{\phi}}}_{k}(\tau) &=  \tilde{\boldsymbol{\rho}} [\boldsymbol{K}_{\phi}(\boldsymbol{t},\boldsymbol{t})+\sigma_{\phi}^2\boldsymbol{I}]^{-\top} \boldsymbol{k}_{\phi}(\boldsymbol{t}, \tau),
		\\
		\tilde{\boldsymbol{a}}_{k}(\tau)  &= \tilde{\boldsymbol{\alpha}} [\boldsymbol{K}_a(\boldsymbol{t},\boldsymbol{t})+\sigma_a^2\boldsymbol{I}]^{-\top} \boldsymbol{k}_a(\boldsymbol{t}, \tau),
	\end{aligned}
\end{equation}
where $\boldsymbol{t} = [t_{1}, t_{2}, \cdots, t_{N}]$ is timestamp vector of desired latent states, $\boldsymbol{K}(\boldsymbol{t}, \boldsymbol{t}) = k(\boldsymbol{t}, \boldsymbol{t})$ is $N \times N$, $\boldsymbol{k}(\boldsymbol{t}, \tau) = k(\boldsymbol{t}, \tau)$ is $N \times 1$,  $\tilde{\boldsymbol{\rho}} = [\tilde{\boldsymbol{\rho}}_{1}, \tilde{\boldsymbol{\rho}}_{2}, \cdots, \tilde{\boldsymbol{\rho}}_{N}]$ is $3 \times N$, and $\tilde{\boldsymbol{\alpha}} = [\tilde{\boldsymbol{\alpha}}_{1}, \tilde{\boldsymbol{\alpha}}_{2}, \cdots, \tilde{\boldsymbol{\alpha}}_{N}]$ is $3 \times N$.
According to \cite{le2023continuous}, the latent states can be recovered by means of constructing several optimization problems where the latent states $\tilde{\boldsymbol{\rho}}$ and $\tilde{\boldsymbol{\alpha}}$ in \eqref{eq:posteriorGP} are optimized to predict a bunch of true IMU measurements as accurate as possible.
After that, the preintegrated velocity $\Delta \tilde{\boldsymbol{\nu}}_{k}(\tau)$, position $\Delta \tilde{\boldsymbol{r}}_{k}(\tau)$ and rotation vector increment $\Delta \tilde{\boldsymbol{\phi}}_{k}(\tau)$ can be written as:
\begin{equation}
	\label{eq:Query UGPM}
	\begin{aligned}
		\Delta \tilde{\boldsymbol{\phi}}_{k}(\tau) &= \int_{t_{k}}^{\tau} \tilde{\dot{\boldsymbol{\phi}}}_{k}(t) dt,
		\\
		\Delta \tilde{\boldsymbol{\nu}}_{k}(\tau) &= \int_{t_{k}}^{\tau} \tilde{\boldsymbol{a}}_{k}(t) dt,
		\\
		\Delta \tilde{\boldsymbol{r}}_{k}(\tau) &= \int_{t_{k}}^{\tau} \int_{t_{k}}^{t} \tilde{\boldsymbol{a}}_{k}(s) ds dt,
	\end{aligned}
\end{equation}
where $\tau > t_{k}$ is the queried timestamp. From \eqref{eq:posteriorGP}, we know that the related variables of integration operations in \eqref{eq:Query UGPM} are only $\boldsymbol{k}_{\phi}(\boldsymbol{t}, \tau)$ and $\boldsymbol{k}_a(\boldsymbol{t}, \tau)$. Indeed, the integration on latent states can be calculated analytically when $\boldsymbol{k}_{\phi}(\boldsymbol{t}, \tau)$ and $\boldsymbol{k}_a(\boldsymbol{t}, \tau)$ are given. The efficient analytical preintegration offers an ability for inferring on asynchronous measurements and continuous-time trajectories.
The preintegration for the rotation matrix can be calculated with \eqref{eq:localgppval} as $\Delta \tilde{\boldsymbol{C}}_{k}(\tau) = \exp(\Delta \tilde{\boldsymbol{\phi}}_{k}(\tau)^{\land})$.  
Also, it is convenient to recover the kinematic state $\boldsymbol{x}(\tau)$ by means of composing the previous state $\boldsymbol{x}(t_{k})$ with \eqref{eq:Query UGPM}. With these composed states, the asynchronous visual projection can be done on arbitrary timestamps.                                             

\subsection{Residual Terms for GP-Preintegration}
According to the inertial preintegration method, the residual terms for GPP are built with the integration increments in \eqref{eq:Query UGPM} as follows:
\begin{equation}
	\label{eq:IMU preinte res}
	\begin{aligned}
		\boldsymbol{e}_{\phi} &= \log(\Delta \tilde{\boldsymbol{C}}_{k,k+1}^{\top} \boldsymbol{C}_{k,k+1})^\vee,
		\\
		\boldsymbol{e}_{\nu} &=  \boldsymbol{C}_{k,k+1} \boldsymbol{\nu}_{k+1}-\boldsymbol{\nu}_{k}-\boldsymbol{C}_{k}^{\top}\boldsymbol{g}_{i} \Delta t_{k,k+1}  - \Delta \tilde{\boldsymbol{\nu}}_{k,k+1},
		\\
		\boldsymbol{e}_{r} &= \boldsymbol{C}_{k}^{\top} (\boldsymbol{r}_{k+1} - \boldsymbol{r}_{k} 
		- \frac{1}{2}\boldsymbol{g}_{i} \Delta t_{k,k+1}^2 ) - \boldsymbol{\nu}_{k}\Delta t_{k,k+1} - \Delta \tilde{\boldsymbol{r}}_{k,k+1},
	\end{aligned}
\end{equation} 
where we have $\boldsymbol{r}_{k} = \boldsymbol{r}_{i}^{bi}(t_{k})$, and $\boldsymbol{\nu}_{k} = \boldsymbol{\nu}_{b}^{bi}(t_{k})$, and $\Delta \tilde{\boldsymbol{\nu}}_{k,k+1} = \Delta \tilde{\boldsymbol{\nu}}_{k}(t_{k+1})$, and $\boldsymbol{C}_{k} = \boldsymbol{C}_{ib}(t_{k})$, and $\boldsymbol{C}_{k,k+1} = \boldsymbol{C}_{k}^{\top} \boldsymbol{C}_{k+1}$, and so on. All notations and frames of \eqref{eq:IMU preinte res} follow the definitions in Sec.~\ref{sec:gpse3}.

\section{Event-driven VIO} \label{sec:eVIO}

\begin{figure}[!t]
	\centering
	\includegraphics[width=3.2in]{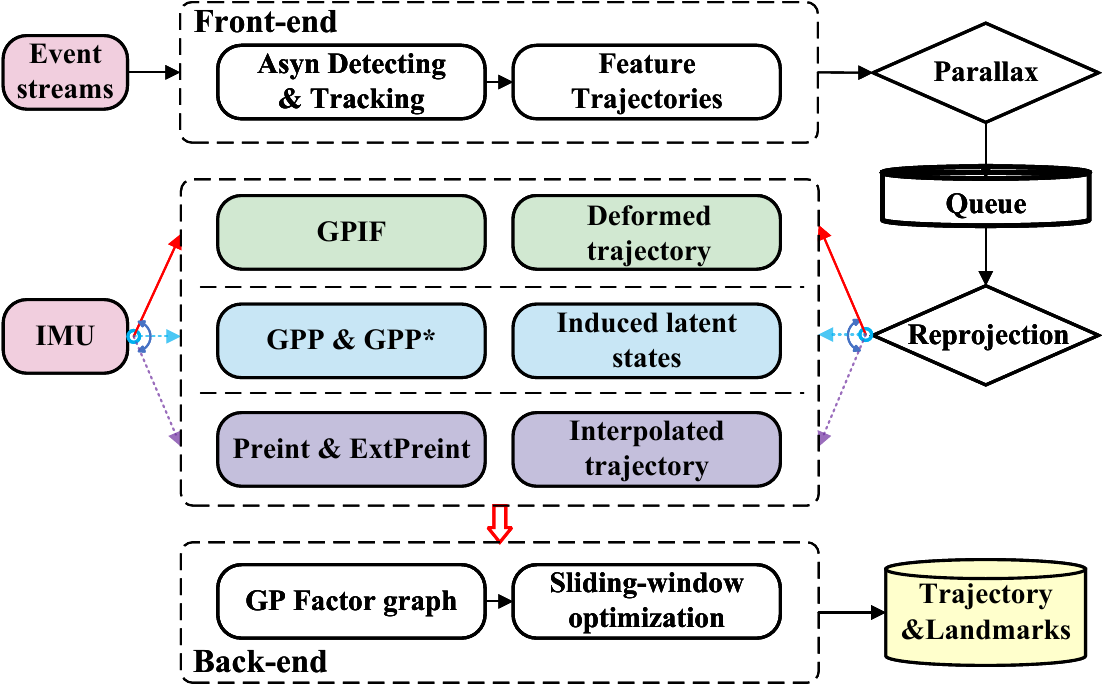}
	\caption{System framework of AsynEIO.}
	\label{fig:sys_framework}
\end{figure}

Given the GPIF and GPP, we design an event-inertial odometry that can estimate motion trajectory from pure event streams and inertial measurements.  
Both GPIF and GPP, in conjunction with their variants, are built into the same EIO system (as shown in Fig.~\ref{fig:sys_framework}). 
This design allows us to switch to different inertial fusion methods and compare them fairly. 

\begin{figure}[!t]
	\centering
	\includegraphics[width=3.2in]{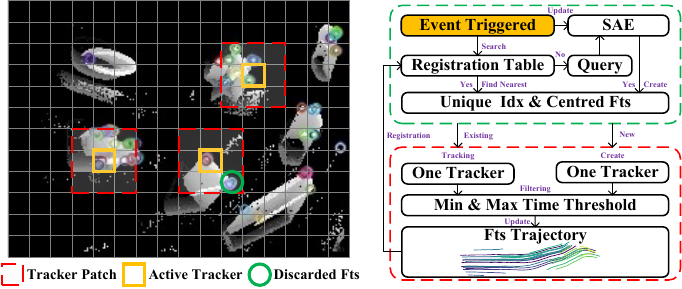}
	\caption{Event-driven feature trajectory generation and management. The registration table (left) marks the active pixels patches, which limits the detection and tracking regions and enables an event-driven front-end (right). All potential event features (labeled as Discarded Fts) within the active Tracker Patch will be abandoned. }
	\label{fig:frontend1}
\end{figure}

\subsection{Event-Driven Front-end}

\begin{algorithm}[!t]
	\caption{Event-Driven Visual Front-end}
	\label{alg AsynEIO_Frontend}
	\KwData{	Event stream queue $\mathcal{E} = \langle  \boldsymbol{e}_1,\boldsymbol{e}_2,...,\boldsymbol{e}_n \rangle$, \\
		\quad \quad \quad Pair of fts trajectory and tracker $\mathcal{P}_{i} = (\boldsymbol{F}_{i}, \mathcal{T}_{i})$ \\
		\quad \quad \quad Map function $\mathcal{M}: i \to \mathcal{P}_{i}$, \\
		\quad \quad \quad Registration table and SAE $\mathcal{G},\mathcal{S} \in \mathbb{R}^{w,h}$,\\
		\quad \quad \quad Set of pairs  $\Phi = \{\mathcal{P}_{i} | i \in \mathcal{M}\}$.}
	\KwResult{Feature trajectory set  $\boldsymbol{F} = \{ \boldsymbol{F}_{i} | i \in \mathcal{M} \}$.}
	$i \leftarrow 0$, and $\Phi, \mathcal{M}, \boldsymbol{F}  \leftarrow \emptyset$,  and $\mathcal{G} \leftarrow -1$, and $\mathcal{S} = 0$\;
	\While{$\mathcal{E}$ not empty}{
		$\boldsymbol{e}_{j}$ = $\mathcal{E}$.pop()\;
		$\mathcal{S}$.update($\boldsymbol{e}_{j}$)\;
		\If{$\mathcal{G}$.Search($\boldsymbol{e}_{j}$)}{
			$k \leftarrow \mathcal{G}$.nearest($\boldsymbol{e}_{j}$)\;
			$\mathcal{P}_{k} \leftarrow \mathcal{M}(k)$\;
			Tracking: Update tracker $\mathcal{T}_{k}$ with $\boldsymbol{e}_{j}$\;
			\If{ $t_{min} < \boldsymbol{e}_{j}.t - \mathcal{T}_{k}.time < t_{max}$ }{
				$\mathcal{T}_{k}.time \leftarrow \boldsymbol{e}_{j}.t$\;
				$\boldsymbol{F}_{k}$.append($\langle \mathcal{T}_{k}.time, \mathcal{T}_{k}.location \rangle$)\;
				Move $k$ to $ \mathcal{T}_{k}.location$ in $\mathcal{G}$\;
			} \ElseIf{$\boldsymbol{e}_{j}.t - \mathcal{T}_{k}.time \geq t_{max}$}{
				Set $-1$ in $\mathcal{G}$ at $\mathcal{T}_{k}.location$\;
				Delete $k$, $\mathcal{P}_k$, $\boldsymbol{F}_k$ in $\mathcal{M}$, $\Phi$ and $\boldsymbol{F}$\;	
				
			}
		} \ElseIf{$len(\boldsymbol{F})  < fts_{max} $}{
			\If{$\mathcal{S}$.Query($\boldsymbol{e}_{j}$) is FA-Harris}{
				$i \leftarrow i+1$\;
				Create $\boldsymbol{F}_{i} = \langle \boldsymbol{e}_{j}.t, \boldsymbol{e}_{j}.location  \rangle$\;
				Create a tracker $\mathcal{T}_{i}$, Update with $\boldsymbol{e}_{j}$\;
				$\mathcal{T}_{i}.time \leftarrow \boldsymbol{e}_{j}.t$\;
				Create $\mathcal{P}_{i}$ with $\boldsymbol{F}_{i}$ and $\mathcal{T}_{i}$\;
				Add $i\rightarrow \mathcal{P}_{i}$ to $\mathcal{M}$, $\boldsymbol{F}_{i}$ to $\boldsymbol{F}$, $\mathcal{P}_{i}$ to $\Phi$\;
				Registration: $\mathcal{G}$.record($i$) at $\boldsymbol{e}_{j}.location$\;
			}
		}
	}
\end{algorithm}

To realize high-temporal resolution, we design an asynchronous event-driven front-end where sparse features are detected  and tracked in an event-by-event manner, as depicted in Fig.~\ref{fig:frontend1}. Although separate event-driven feature detecting and tracking methods have been proposed in previous literature, a whole event-driven front-end needs further design on strategies of screening and maintenance. 

\begin{figure}[!t]
	\centering
	\includegraphics[width=3.0in]{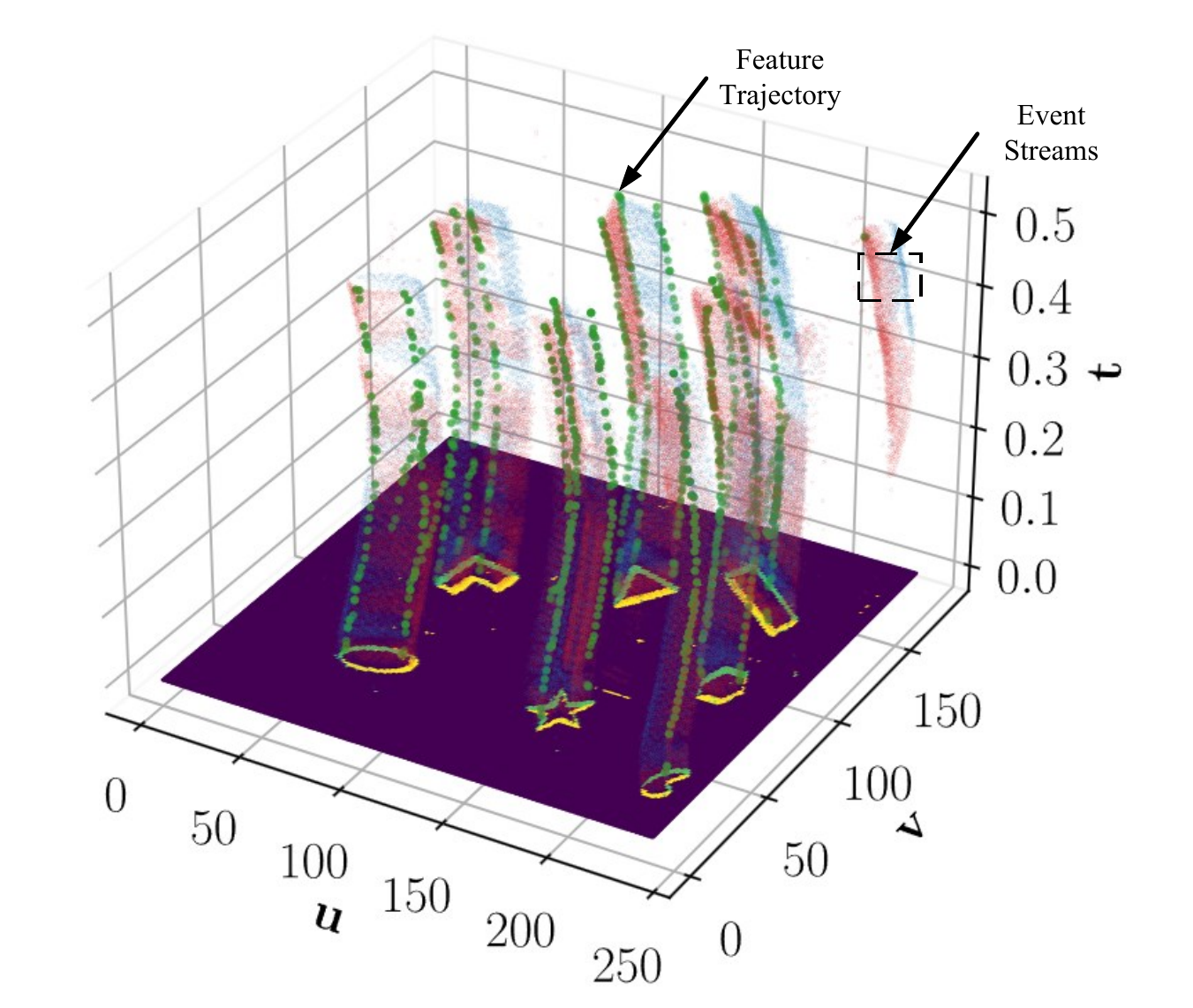}
	\caption[Illustration of feature trajectories.]{Illustration of event streams and feature trajectories. When the event camera observes some landmarks and moves in the scenario, the event stream triggered by the same landmark will be managed as a feature trajectory.}
	\label{feature trajectory illustration}
\end{figure}

In AsynEIO, The event-driven front-end is abstracted as several basic 
operators: 1) \emph{Search}, 2) \emph{Query}, 3) \emph{Registration}, 4) \emph{Tracking}. 
Each reported event first invokes the \emph{Search} operator to determine whether there 
already exists an active tracker within its neighbor (labeled as orange solid boxes in 
Fig.~\ref{fig:frontend1}). The \emph{Query} operator try to detect a new feature within 
the inactive zone, while the \emph{Tracking} operator feed the reported event to the active 
tracker to update the feature trajectory (as visualized in Fig.~\ref{feature trajectory illustration}). New created trackers and feature trajectories are 
recorded using the \emph{Registration} operator. All events located within the active Tracker 
Patch (labeled as red dashed boxes) will not be queried, even if they are exact corner features, 
such as regions in green circles of Fig.~\ref{fig:frontend1}.  A registration table is designed 
to ensure that all operators only access or modify within small patches. Each active tracker is 
triggered by only a small fraction of events (within the Active Tracker). The event coordinates 
strictly constrain the pixel area of \emph{Query} operators, which is more efficient than conventional feature detection in intensity images.

As shown in Alg.~\ref{alg AsynEIO_Frontend}, the registration table $\mathcal{G}$ records the unique identifier $i$ of new event features. The pixel location of each event feature can be infer from its index in the registration table. Then, the unique identifier is mapped to the corresponding feature trajectory $\boldsymbol{F}_{i}$ and its independent tracker $\mathcal{T}_{i}$ (combined into a pair variable $\mathcal{P}_{i}$). When an event measurement $\boldsymbol{e}_{j}$ is obtained, we first update the Surface of Active Events (SAE) $\mathcal{S}$ and \emph{Search} the registration table $\mathcal{G}$ to determine whether active event features already exist in its neighboring region (Line 4-5). If there is at least one active event feature, we utilize the identifier-tracker map $\mathcal{M}$ to find the corresponding tracker and feed the event measurement to it (Line 6-8). The movement of active feature will be reported by the tracker and collected into the feature trajectory (Line 11). The searched unique identifier $k$ in the registration table will also be moved to the corresponding location (Line 12). Otherwise, we \emph{Query} the SAE $\mathcal{S}$ to determine if a FA-Harris \cite{li2019fa} corner feature can be detected at the pixel location of $\boldsymbol{e}_{j}$ (Line 16-17). Once a new event feature is reported, an individual tracker $\mathcal{T}_{i}$ \cite{alzugaray2020haste} is created and added in the set $\Phi$ (Line 18-23) for subsequent tracking. Accordingly, it is registered in the registration table $\mathcal{G}$ (Line 24).  

Since the active tracker is triggered by asynchronous events, the inactive trackers should be removed promptly. To realize it, we assume the active tracker should be triggered frequently within a maximum time threshold. We periodically delete active features that have not been updated beyond the maximum time threshold (Line 13-15). In addition, we also set a minimum time threshold to avoid the feature trajectory collecting a mass of redundant measurements (Line 9-12). With the registration maintenance and time-threshold screening mechanisms, the result front-end can obtain sparse and asynchronous feature tracking.

\subsection{Asynchronous Repojection}

\begin{figure}[!t]
	\centering
	\includegraphics[width=3.0in]{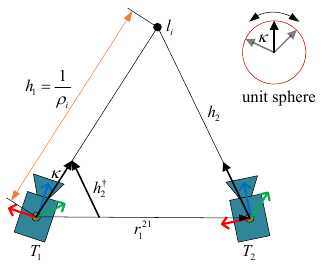}
	\caption[Inverse depth parametrization.]{Inverse depth parametrization.}
	\label{project_illu}
\end{figure} 

The feature trajectory, uniquely indexed by $i$, can be defined as
\begin{equation}
	\label{ft}
	\boldsymbol{F}_{i} \triangleq \langle  \boldsymbol{m}_1^i,\boldsymbol{m}_2^i,...,\boldsymbol{m}_n^i \rangle ,
\end{equation}
which represents an ordered spatial-temporal association set. The $\boldsymbol{m}_{j}^i$ indicates the $j$-th event feature measurement of the landmark $\boldsymbol{l}_i$ at timestamp $t_{i,j}$, and it  contains 
\begin{equation}
	\label{eq:fts_pixel_position}
	\boldsymbol{m}^{i}_{j} \triangleq \{ t_{i, j}, \tilde{\boldsymbol{q}}_{i,j} \},
\end{equation}
where $t_{i, j}$ is the measurement timestamp and $\tilde{\boldsymbol{q}}_{i,j} = [x_{i, j}, y_{i, j}]^{\top}$ is the observed pixel position.

When a feature trajectory has collected enough measurements for triangulation, 
the induced trajectory will be utilized to triangulate and embed these measurements into the underlying factor graph. 
Given the triangulated landmark $\boldsymbol{l}_{i} = [x_{i}, y_{i}, z_{i}]^{\top}$ and the event feature measurement $\boldsymbol{m}_{j}^{i}$ as defined in \eqref{eq:fts_pixel_position},  the projection model can be given by
\begin{equation}
	\label{eq:visual_residual}
	\boldsymbol{e}_{c} (t_{i,j})= \tilde{\boldsymbol{q}}_{i,j} - \frac{1}{d_{i}} \boldsymbol{K} (\boldsymbol{T}(t_{i,j}) \boldsymbol{T}_{bc})^{\top} \boldsymbol{l}_{i},
\end{equation}
where $\boldsymbol{K}$ indicates the intrinsic matrix, 
$d_{i}$ is the depth of this landmark in current camera frame, 
$\boldsymbol{T}_{bc}$ represents the extrinsic transformation matrix 
between the event camera and IMU. 
The 
kinematic state $\boldsymbol{T}(t_{i,j})$ can be queried 
using \eqref{eq:Tlocal} and \eqref{eq:query_lin_state} for the GPIF scheme. 
In contrast, the GPP can analytically infer $\boldsymbol{T}(t_{i,j})$ from 
induced latent states using linear operators, as shown in \eqref{eq:localgppval} and \eqref{eq:Query UGPM}.

The inverse depth parametrization \cite{civera2008inverse} is adopted to enhance the numerical stability and realize undelayed initialization. 
Simplify \eqref{eq:visual_residual} as
\begin{equation}
	\boldsymbol{e}_c  = \boldsymbol{q}_{2} - \tilde{\boldsymbol{q}}_{2}.
\end{equation}
Define the inverse depth $\rho$ and the directional vector $\boldsymbol{\kappa}$ in unit sphere of the first observed camera pose $\boldsymbol{T}_{1}$ to represent the landmark $\boldsymbol{l}_{i}$. As displayed in Fig.~\ref{project_illu}, the latter camera pose $\boldsymbol{T}_{2}$ should observe the ray $\boldsymbol{h}_{2}$ and have a reprojected pixel location $\boldsymbol{q}_{2}$. When the poses and landmark are estimated precisely, this reprojected pixel location should coincide with the visual tracking result $\tilde{\boldsymbol{q}}_{2}$ at the same camera pose. According to the geometric constraints in Fig.~\ref{project_illu}, the projected directional vector can be given by
\begin{equation}
	\boldsymbol{h}_{2}^{\dagger} = \boldsymbol{C}_{12}^{\top} (\boldsymbol{\kappa} -\rho \boldsymbol{r}_{1}^{21}) 
	=\boldsymbol{P} \boldsymbol{T}_{12}^{-1}  \begin{bmatrix}
		\boldsymbol{\kappa} \\
		\rho
	\end{bmatrix} 
	=\begin{bmatrix}
		h_{x} \\
		h_{y} \\
		h_{z}
	\end{bmatrix},
\end{equation}
where $\boldsymbol{P} = \begin{bmatrix}
	\boldsymbol{1}^{3\times 3}, \boldsymbol{0}^{3\times 1} 
\end{bmatrix}$.
Therefore, the projection function can be written as 
\begin{equation}
	\boldsymbol{q}_{2} = \boldsymbol{K} (\boldsymbol{h}_{2}^{\dagger})^{\flat},
\end{equation}
where $(\boldsymbol{h}_{2}^{\dagger})^{\flat} = [h_{x}  / h_{z}, h_{y} / h_{z}, 1]^{\top}$.
The infinity landmark or the pure rotation can be represented by zero inverse depth $\rho=0$ for all time. In both cases, the pose and inverse depth can be optimized unambiguously, allowing for the immediate initialization of feature trajectories. Although inverse depth offers immediate initialization, it introduces additional state variables (6-DoF for the first observed pose) for each projection factor. When the first observed pose occurs between samples of the GP trajectory, the complexity increases further due to GP interpolation, involving two adjacent trajectory points with a total of 36 dimensions.  To mitigate this issue, we interpolate along the feature trajectory to identify the nearest sampling timestamp on the motion trajectory for computing $\boldsymbol{\kappa}$. This process initializes each feature trajectory for inverse depth optimization and reduces 30 dimensions for each projection factor. Benefited from high-time resolution of visual front-end, the interpolation on feature trajectories normally has negligible deviations.

\subsection{Overall Cost Function in Sliding-Window}

\begin{figure}[!t]
	\centering
	\includegraphics[width=3.2in]{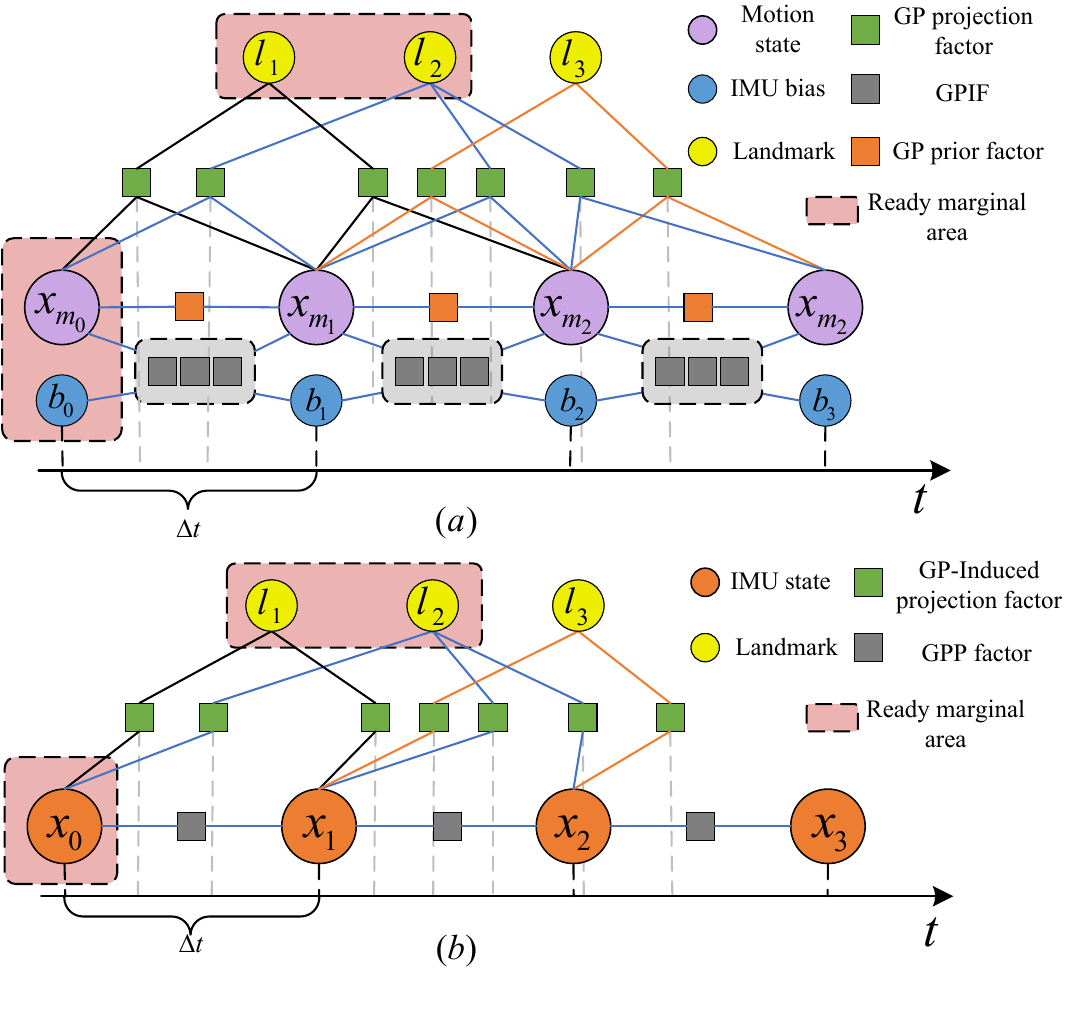}
	\caption{Factor graph of two main asynchronous fusing schemes. (a) Factor graph for GPIF, where asynchronous projection is achieved through GP interpolation on the motion trajectory. (b) Factor graph for GPP, which enables analytical preintegration at arbitrary timestamps to support asynchronous projection. The dynamic marginalization and sliding-window optimization are adopted to reduce the computational consumption.}
	\label{factor_graph}
\end{figure}

The key innovation of GPIF lies in its ability to handle asynchronous IMU measurements while eliminating the need for preintegration operations. For the Micro-Electro-Mechanical Systems (MEMS) IMU, the preintegration would introduce collective errors and deteriorate the estimation results after a fewer second. Instead, the GPIF establishes detailed relationships between the differential SE(3) states of kinematic trajectories and the physical measurements obtained from  IMUs. The residuals, Jacobians, and covariance matrices for GPIF can be elegantly formulated and seamlessly integrated into GP-based factor graphs. The overall cost for our event-driven VIO with GPIF is given by 
\begin{align}
	\label{eq:OP}
	\mathop{\min}_{\boldsymbol{x}, \boldsymbol{l}} \underbrace{\sum \Vert \boldsymbol{e}_{k} \Vert_{\boldsymbol{Q}_{k+1}}^2}_{\boldsymbol{WNOJ\ prior}} + \underbrace{\sum \Vert \boldsymbol{e}_{bg}(t_k)  \Vert_{\boldsymbol{Q}_{bg}}^{2} + \sum \Vert \boldsymbol{e}_{bg}(t_k)  \Vert_{\boldsymbol{Q}_{bg}}^{2}}_{\boldsymbol{bias\ prior}} \nonumber\\
	+ \underbrace{\sum \Vert \boldsymbol{e}_{c}(t_{i,j}) \Vert_{\boldsymbol{R}}^2}_{\boldsymbol{event\ cost}}+ 
	\underbrace{\sum \Vert \boldsymbol{e}_{g}(\tau)  \Vert_{\boldsymbol{Q}_{g}}^{2} + \sum \Vert \boldsymbol{e}_{a}(\tau)  \Vert_{\boldsymbol{Q}_{a}}^{2}}_{\boldsymbol{GPIF\ cost}}, 
\end{align}
where $\boldsymbol{R}, \boldsymbol{Q}$ are the covariance matrices, the WNOJ prior $\boldsymbol{e}_{k}$ is defined as \eqref{PriorRes}, the bias prior is expressed in \eqref{eq:bias_prior}, the event cost is denoted in \eqref{eq:visual_residual}, and the GPIF cost is given by \eqref{eq:inertial_residual}. The estimated states of \eqref{eq:OP} include kinematic states $\boldsymbol{x}$ and landmarks $\boldsymbol{l}$ in the sliding-window. Specifically, the kinematic states $\boldsymbol{x}$ for GPIF consist of GP trajectory points $\boldsymbol{x}(t_{k})$ and IMU biases \{$\boldsymbol{b}_{a}(t_{k}), \boldsymbol{b}_{g}(t_{k})$\}, as defined in \eqref{eq:kin_state} and \eqref{eq:acc_measure}. In addition, the event cost and GPIF cost in \eqref{eq:OP} leverage the interpolation formulas \eqref{eq:query_lin_state} and \eqref{eq:Tlocal}-\eqref{eq:localstate} to conduct asynchronous residuals.
Compared with preintegration schemes, the inertial cost of GPIF directly applies the measurement covariance (i.e., $\boldsymbol{Q}_{g}$ and $\boldsymbol{Q}_{a}$), which avoids the computational consumption paid for the covariance propagation. Moreover, this propagation becomes unreliable when dealing with partial loss or asynchronous inertial measurements. Similarly, the summary cost for GPP can be defined as
\begin{align}
	\label{eq:OP_gpp}
	\mathop{\min}_{\boldsymbol{x}, \boldsymbol{l}} \underbrace{\sum \Vert \boldsymbol{e}_{bg}(t_k)  \Vert_{\boldsymbol{Q}_{bg}}^{2} + \sum \Vert \boldsymbol{e}_{bg}(t_k)  \Vert_{\boldsymbol{Q}_{bg}}^{2}}_{\boldsymbol{bias\ prior}} + \underbrace{\sum \Vert \boldsymbol{e}_{c}(t_{i,j}) \Vert_{\boldsymbol{R}}^2}_{\boldsymbol{event\ cost}}  \nonumber\\
	+ \underbrace{\sum \Vert \boldsymbol{e}_{\phi}(t_{k})  \Vert_{\boldsymbol{Q}_{\phi}}^{2} + \sum \Vert \boldsymbol{e}_{\nu}(t_{k})  \Vert_{\boldsymbol{Q}_{\nu}}^{2} + \sum \Vert \boldsymbol{e}_{r}(t_{k})  \Vert_{\boldsymbol{Q}_{r}}^{2}}_{\boldsymbol{GPP\ cost}}, 
\end{align}
in which the bias prior and event cost terms are the same as GPIF, and the GPP cost is defined as \eqref{eq:IMU preinte res}. Different from GPIF, The GPP has lower-dimension kinematic states including \{$\boldsymbol{T}(t_{k}), \boldsymbol{\nu}(t_{k}), \boldsymbol{b}_{a}(t_{k}), \boldsymbol{b}_{g}(t_{k})$\}, where the interpolated state $\boldsymbol{x}(\tau)$ for asynchronous event residuals can be inferred from induced latent states and linear operators, as done in \eqref{eq:localgppval} and \eqref{eq:Query UGPM}.

\begin{figure}[!t]
	\centering
	\includegraphics[width=3.0in]{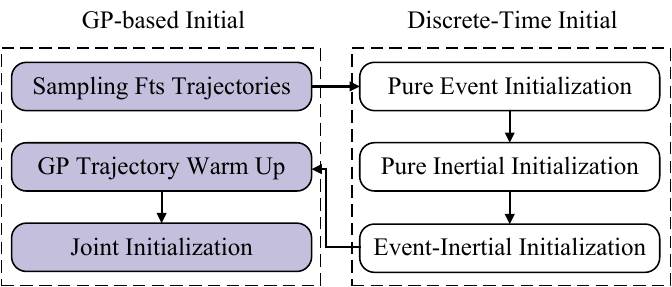}
	\caption{Event-inertial initialization in AsynEIO.}
	\label{event_inertial_initial}
\end{figure}

\subsection{Event and Inertial Initialization}

We design an initialization scheme (as shown in Fig.~\ref{event_inertial_initial}) to warm up our event-inertial system. The asynchronous continuous-time initialization is first converted to a synchronous discrete-time problem using a sampling strategy. A standard visual-inertial initialization method \cite{qin2018vins} is adopted  to achieve initial estimating of pose, linear velocity and bias. With these known states, the remaining unknown states on the continuous-time GP trajectory are solved by a warm up optimization as  
\begin{equation}
	\mathop{\min}_{\boldsymbol{\omega}, \dot{\boldsymbol{\varpi}}} \underbrace{\sum \Vert \boldsymbol{e}_{k} \Vert_{\boldsymbol{Q}_{k+1}}^2}_{\boldsymbol{WNOJ\ prior}} + \lambda_{1} \Vert\boldsymbol{\omega}_{0}\Vert^{2} + \lambda_{2} \Vert \dot{\boldsymbol{\varpi}}_{0} \Vert^{2},
\end{equation}
where $\boldsymbol{\omega}_{0}$ and $\dot{\boldsymbol{\varpi}}_{0}$ are the rotational velocity and generalized acceleration of the first trajectory point, $\lambda_{1}$ and $\lambda_{2}$ are two weight coefficients. This procedure adds some zero priors and limits the drift of GP trajectory in null space.
Finally, a joint optimization as \eqref{eq:OP} is built to add all asynchronous measurements and refine overall state variables.

\section{Experiments}\label{sec:experiments}

\begin{figure*}[thpb]
	\centering
	\includegraphics[width=6.5in]{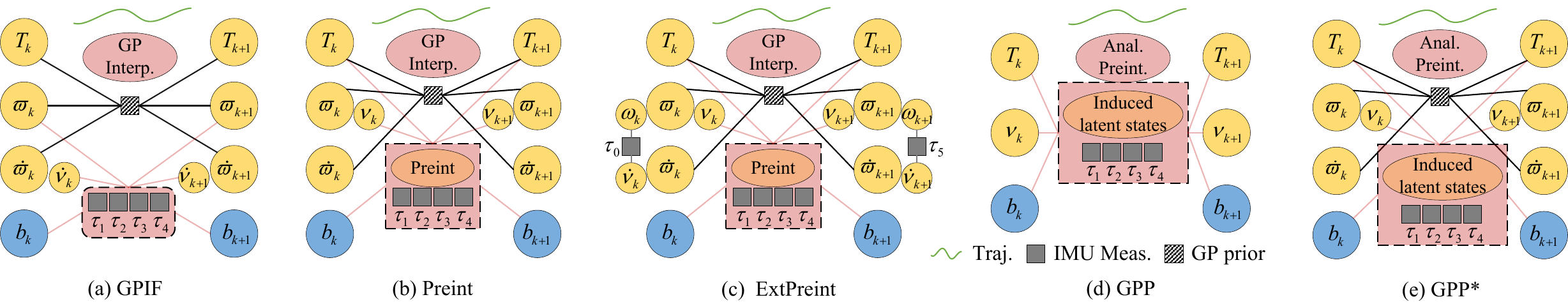}
	\caption{Summary of all inertial factors. Besides the aforementioned three inertial factors (a, b, d), we further extend the preintegration factor as shown in (c). The direct observation \eqref{eq:inertial_residual} about the rotation velocity and body acceleration is built in the factor graph. (a) GPIF, (b) discrete preintegration,  and (c) extended preintegration factor obtain continuous-time trajectory by GP interpolation while the GPP directly induces a series of latent states to conveniently infer an arbitrary inter-state by composing an analytic integration.}
	\label{multi_inertial_factor_graph}
\end{figure*}

We analyse and verify the main modules and the whole pipeline of AsynEIO. Especially, we trend to demonstrate the performances of various inertial fusion schemes for event-based visual-inertial odometry under the continuous-time GP regression framework.
Firstly, the asynchronous front-end and the reprojection rejection modules are showed by tracking and screening real event streams. Secondly, we compare the GP-based continuous trajectories defined in different state spaces when they are applied to describe the rigid motion. Thirdly, we analysis the motion trajectory deformed by various inertial factors and report their ability for offering better trajectory priors. Then, the whole pipeline is comprehensively compared with the state-of-the-art in a mass of challenging scenarios. We report the estimated trajectories and errors. Eventually, the runtime of AsynEIO is countered and analyzed.

\subsection{Implementation Details}

All inertial fusion schemes are defined as individual factors using GTSAM\footnote{https://github.com/borglab/gtsam}.  
The asynchronous reprojection factors and marginalization are realized based on the previous work \cite{li2024asynchronous}.
Different inertial factors are integrated into the back-end factor graph to facilitate switching comparison experiments.
The event-driven front-end is developed based on Alg.~\ref{alg AsynEIO_Frontend}, where FA-Harris \cite{li2019fa} and HASTE \cite{alzugaray2020haste} are modified to detect and track event features, respectively.
The whole system is implemented in C++ and utilizes ROS Noetic on Ubuntu 20.04 LTS for data subscription and publishing. The average runtimes are obtained using an Intel Xeon Gold 6248R@4GHz and a single NVIDIA A6000 GPU.

\begin{table}[!t]
	\centering
	\renewcommand{\arraystretch}{1.5}
	\caption{Configurations of Different Fusion Methods}
	\label{fig:inertial_factors_config}
	\resizebox{3.4in}{!}{
		\begin{threeparttable}
			\begin{tabular}{ l | ccc}
				\toprule[1.5pt]
				Methods 		& Inertial Cost & Motion Prior	& Interpolation \& Projection \\
				\midrule
				GPIF	 		& Raw IMU Data		& WNOJ		& GP Trajectory\\	
				Preint 			& Preintegration	& WNOJ 		& GP Trajectory\\
				ExtPreint 	 	& Preintegration \& Raw IMU Data & WNOJ & GP Trajectory\\
				GPP				& GP-Preintegration	& None 		& Latent State \& Linear Operator\\
				$\text{GPP}^{\ast}$& GP-Preintegration		& WNOJ & Latent State \& Linear Operator\\
				\bottomrule[1.5pt]
			\end{tabular}
			$\text{GPP}^{\ast}$ is a variant of GPP, and ExtPreint can be regarded as a synthesis of Preint and GPIF. 
		\end{threeparttable}
	}
\end{table}

For comprehensive comparisons, we implement five different inertial fusion schemes, as listed in Table~\ref{fig:inertial_factors_config}. Among them, the GPIF and GPP are two main comparison categories. The ExtPreint is a synthetic of Preint and GPIF and the $\text{GPP}^{\ast}$ is a variant of the original GPP.
The GPIF in Fig.~\ref{multi_inertial_factor_graph}(a) can directly build inertial residual errors on the continuous-time GP trajectory using raw IMU measurements (as expressed in \eqref{eq:OP}). 
The discrete preintegration (Preint) firstly integrates raw IMU measurements into relative observations, then constrains the adjacent GP trajectory points (as illustrated in Fig.~\ref{multi_inertial_factor_graph}(b)). We further extend the discrete preintegration using the thought of GPIF. The raw measurements close to the GP trajectory points are leveraged to conduct direct observations using \eqref{eq:inertial_residual}. In conjunction with the preintegration observations, the extended preintegration factor (ExtPreint) appears to represent a trade-off between instantaneous and cumulative observations (as displayed in Fig.~\ref{multi_inertial_factor_graph}(c)). Unlike the aforementioned schemes, the GPP in Fig.~\ref{multi_inertial_factor_graph}(d) builds inertial cost using preintegration results and enables asynchronous reprojection by interpolating on latent states, as represented in \eqref{eq:OP_gpp}. 
The $\text{GPP}^{\ast}$ in Fig.~\ref{multi_inertial_factor_graph}(e) mostly maintains the original GPP, except for the addition of WNOJ priors.

The minimize updating period of feature trajectories is set to $0.002\ s$ for avoiding redundancy observation and guaranteeing the resolution. The camera projection noise is set to $0.8$ and a Cauchy robust kernel function is utilized to mitigate the influence of outliers.
In all evaluations, the latent state frequency $f=400\ Hz$ is utilized to infer the kernel length scale as $3/f$. We have verified that these parameters converge across all tested sequences. 
The calibrated intrinsic parameters of IMUs are directly employed to GPP and Preint. Then, we manually tune the Qc parameter of Preint for different datasets. Finally, GPIF, GPP* and ExtPreint applied the same Qc and inertial parameters for the sake of fairness. For those sequences with smoother ground truth, such as those in UZH-FPV dataset, Qc is set to 10. In contrast, a larger Qc is more appropriate for aggressive motion (like 150 and 200 for stereo-HKU dataset).

\begin{figure}[!t]
	\centering
	\includegraphics[width=3.2in]{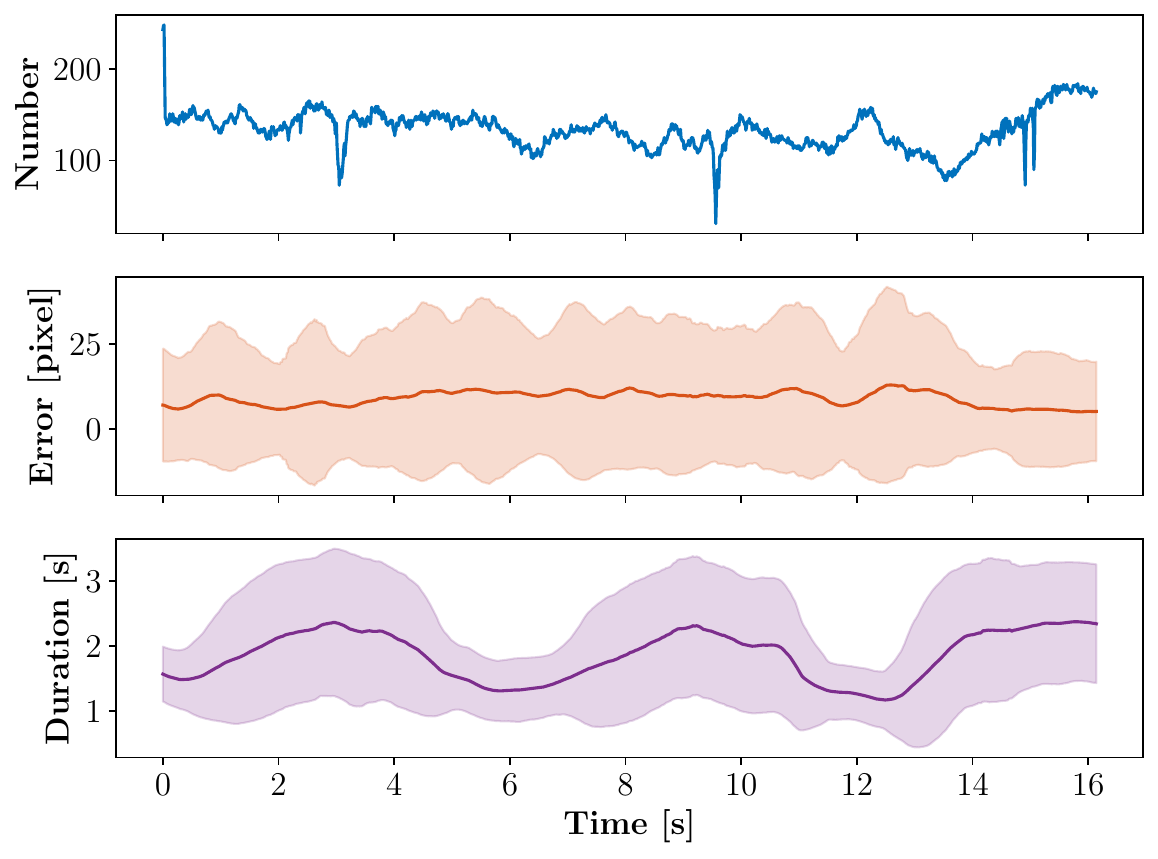}
	\caption{Asynchronous front-end tracking results. The shadow area indicates the standard deviation.}
	\label{asyn_frontend_result}
\end{figure}

\subsection{Asynchronous Tracking Results on Real Event Streams}

\begin{figure}[!t]
	\centering
	\includegraphics[width=3.2in]{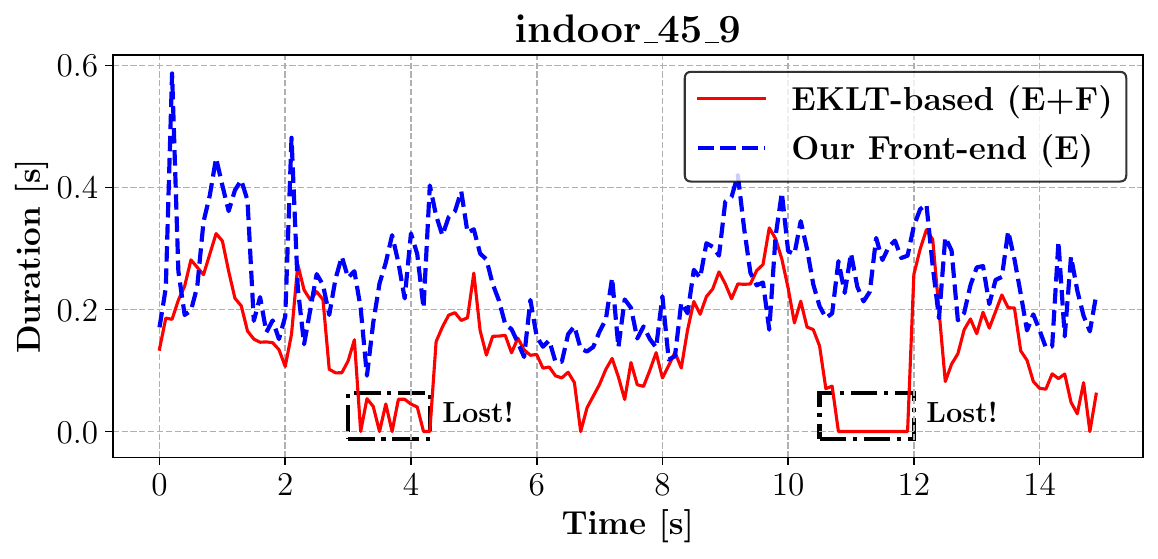}
	\caption{Average duration of tracked feature trajectories with different front-ends.   }
	\label{frontend_indoor_45_9}
\end{figure}

The final estimation results of AsynEIO are tightly influenced by the quality of asynchronous tracking front-end. Ideally, we constantly need ample feature trajectories and low tracking errors all the time. However, it's inevitable to introduce lots of tracking errors and lack enough corner features for the sensor noise and an imperfect tracker. In practice, the triangulation and reprojection rejection can effectively alleviate the mismatch tracking errors. Benefited from the inertial measurements, lacking feature trajectories in some small periods is also acceptable. 
As shown in Fig.~\ref{asyn_frontend_result}, the number of tracked feature trajectories can maintain basic stability. The reprojection errors fluctuate around zero, but have a high standard deviation (shaded area). In practice, we discard those feature trajectories with an error greater than 10 pixels. The duration varies between 1-3 seconds, which basically meets the requirement of the event-inertial odometry.
It is noteworthy that our feature tracking is different from both frame-like and TS-based methods. In AsynEIO, we directly tracking event-based FA-Harris features using raw event streams whenever an individual event is triggered (as in Fig.~\ref{feature trajectory illustration}). On this basis, the raw time-resolution of event camera is maximally preserved. According to the Nyquist Sampling Theorem, the double-frequency is necessary to recover raw signals. As a result, our AsynEIO gain the ability to identify a super high-speed or high-frequency motion. We also compare our asynchronous front-end with the EKLT-based front-end \cite{li2024asynchronous} in terms of the feature trajectories number, reprojection error and duration. As shown in Table~\ref{error_table_tracking_result}, our asynchronous front-end can achieve longer tracking duration than EKLT front-end in all tested sequences. Although our front-end has higher tracking error in basic scenarios,  it generally outperforms the EKLT-based method in high-speed and HDR environments. In fact, the asynchronous mechanism can immediately initialize new feature trajectories when the old trajectories are rejected by reprojection check. On the contrary, the EKLT-based front-end may completely lose all feature trajectories, as displayed in Fig.~\ref{frontend_indoor_45_9}. The final estimation results (labeled as Traj. Error in Table \ref{error_table_tracking_result}) demonstrate that the proposed method significantly improves the success rate and accuracy  in challenging scenarios. 

\begin{table}[!t]
	\centering
	\renewcommand{\arraystretch}{1.2}
	\caption{The Performance Comparison of Different Event Front-ends}
	\label{error_table_tracking_result}
	\resizebox{3.5in}{!}{
		\begin{threeparttable}
			\begin{tabular}{ l  lcccccc cc}
				\toprule[1.5pt]
				\multirow{2}{*}{Scene} &\multirow{2}{*}{Sequence} &\multicolumn{2}{c}{Number}& \multicolumn{2}{c}{Error [pixel]} & \multicolumn{2}{c}{Duration [s]} & \multicolumn{2}{c}{Traj. Error [m]} \\
				\cmidrule(r){3-4}	\cmidrule(r){5-6} \cmidrule(r){7-8} \cmidrule(r){9-10} 
				& & EKLT & Proposed & EKLT & Proposed & EKLT & Proposed & EKLT & Proposed\\
				\cmidrule(r){1-4}	\cmidrule(r){5-6} \cmidrule(r){7-8}  \cmidrule(r){9-10}
				\multirow{2}{*}{Basic} &poster\_6 & \textbf{96.89} & 94.06& \textbf{4.85} &6.07&0.50&\textbf{0.84} & \textbf{0.08} & 0.28\\
				&dynamic\_6 &\textbf{57.59}& 53.67&\textbf{2.87} &5.62&0.78&\textbf{3.47} & 0.10 & \textbf{0.05} \\
				\cmidrule(r){1-4}	\cmidrule(r){5-6} \cmidrule(r){7-8}  \cmidrule(r){9-10}
				\multirow{2}{*}{High-speed} &indoor\_45\_9 & 155.17& \textbf{178.03} &11.38& \textbf{7.86} &0.14&\textbf{0.25} & failed  & \textbf{1.38}\\
				&indoor\_for\_9 &\textbf{129.21} & 107.06&\textbf{3.98} &6.56&0.53&\textbf{0.77} & failed &\textbf{1.53}\\
				\cmidrule(r){1-4}	\cmidrule(r){5-6} \cmidrule(r){7-8} \cmidrule(r){9-10}
				\multirow{2}{*}{HDR} &hdr\_poster & 71.58& \textbf{78.23} &\textbf{5.05}&5.89&1.00& \textbf{1.35} & 0.76 & \textbf{0.14} \\
				&vicon\_hdr &126.95& \textbf{217.64} &18.70&\textbf{8.33}&0.10&\textbf{1.04} & failed & \textbf{0.75} \\
				\bottomrule[1.5pt]
			\end{tabular}
		\end{threeparttable}
	}
\end{table}

\subsection{GP Trajectory Comparison under Different State Spaces}

\begin{figure}[!t]
	\centering
	\includegraphics[width=3.1in]{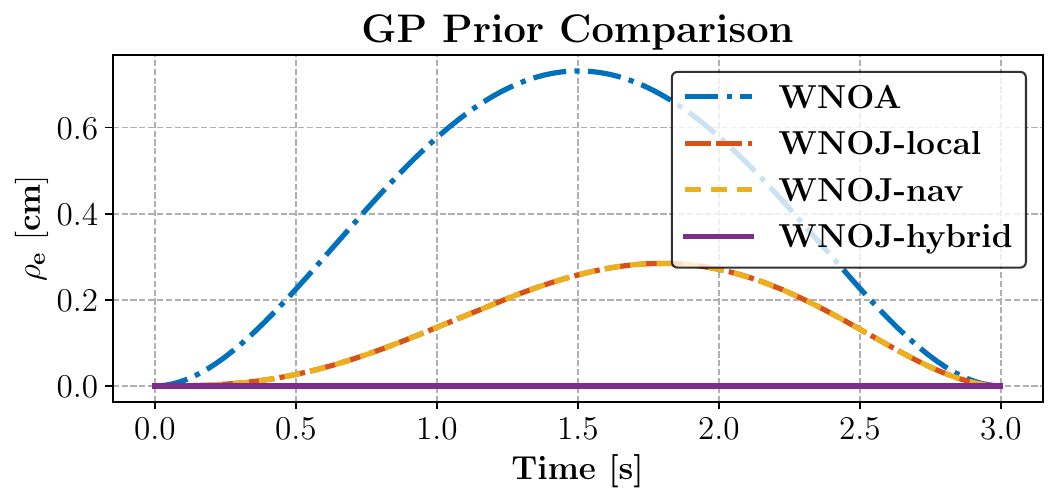}
	\caption{Errors with different GP prior when represent a constant acceleration trajectory.   }
	\label{gp_prior_comparison}
\end{figure}
\begin{figure}[!t]
	\centering
	\includegraphics[width=3.2in]{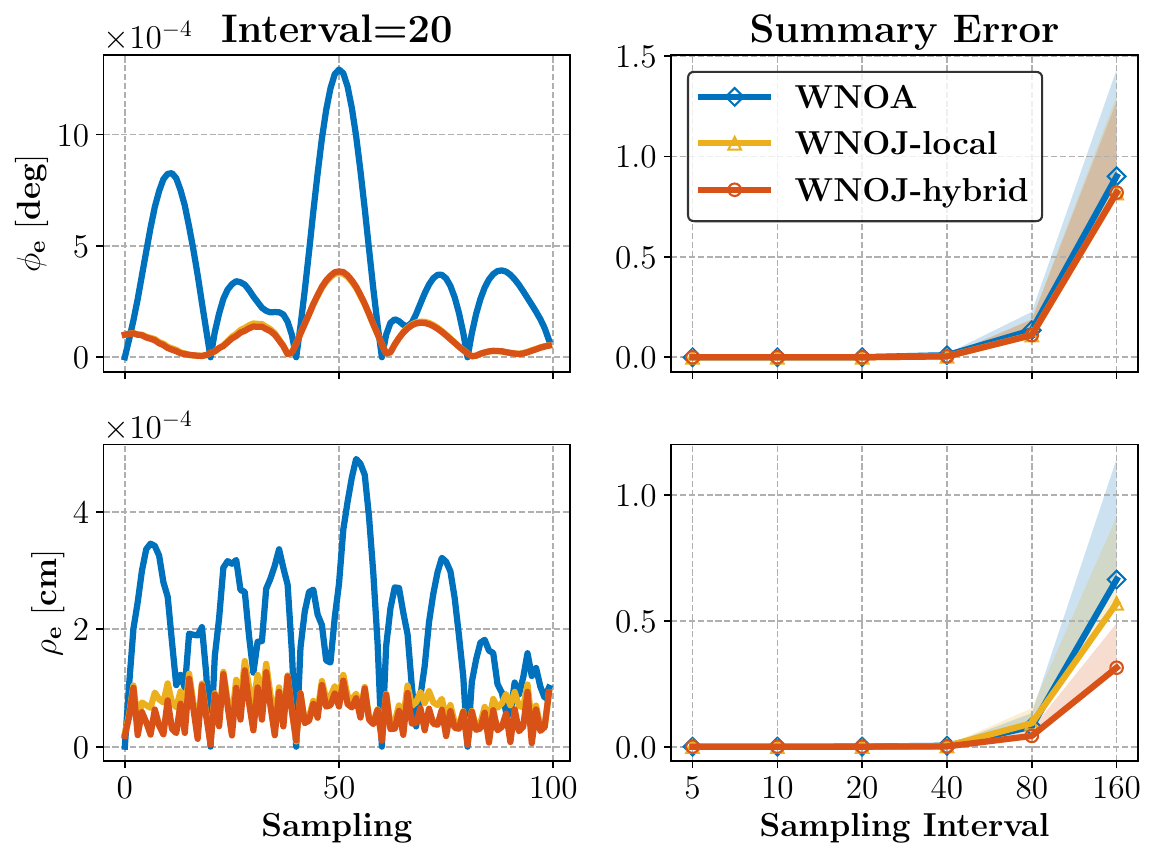}
	\caption{GP trajectory errors of different GP assumptions under various sampling intervals. }
	\label{gp_prior_comparison_fpv}
\end{figure}

Existing state space definitions for GP trajectory mainly include three types: local GP-prior on $SE(3)$, and hybrid GP-prior on $SO(3)$ and $\mathbb{R}^{3}$. In this paper, we mainly adopt the local GP-prior on $SE(3)$. We show that it not only has an elegant form on mathematics but also has a meaningful physical content as derived foregoing. In terms of motion prior levels, the WNOA and WNOJ are two main categories. The WNOA assumes that the acceleration of the motion trajectory obeys the zero-mean Gaussian distribution, while the WNOJ regards the jerk of the motion trajectory as Gaussian noise with zero-means.  The WNOA prior needs a $12$-DoF state space to express the trajectory, and the WNOJ prior needs a $18$-DoF state space. Obviously, the WNOA should achieve higher computational efficiency. Now, we show their expressive ability under different state spaces and GP assumptions. Firstly, a constant acceleration motion trajectory is generated by a nominal kinematics model ($\boldsymbol{\omega}_{i} = [0.13, 0.065, 0.169]\ rad/s$, $\dot{\boldsymbol{\nu}}_{i}^{bi} = [0.5, -0.15, 0.55]\ m/s^{2}$, and $\boldsymbol{\nu}_{i}(t_0) = [1.0,0.0,0.0]$). Then, we fix two endpoints and interpolate internal states under different GP-based assumption and state spaces. The Root-Mean-Square-Error (RMSE) of the trajectory is computed by
\begin{equation}
	\label{gp_traj_error}
	\begin{aligned}
	[\boldsymbol{\phi}_{e}(t), \boldsymbol{\rho}_{e}(t)] = \log(\boldsymbol{T}(t) \boldsymbol{T}^{-1}(t)),  \\ 
	\phi_{e} = \sqrt{ \sum_{k=0}^{n-1} \frac{\Vert \boldsymbol{\phi}_{e}(t_{k}) \Vert^{2}}{n} }, \quad \rho_{e} = \sqrt{  \sum_{k=0}^{n-1} \frac{\Vert \boldsymbol{\rho}_{e}(t_{k}) \Vert^{2}}{n} },
	\end{aligned}
\end{equation}
where $t_{k}$ and $n$ is the timestamp and size of matched poses with the ground truth, respectively.
In our simulation, the rotation error $\phi_{e}(t)$ is zero, so we only consider the translation error $\rho_{e}(t)$. As shown in Fig.~\ref{gp_prior_comparison}, we find that the change of coordinate frame (WNOJ-local and WNOJ-nav) will not affect the GP trajectory. The WNOJ totally has smaller translation error than WNOA when describing a constant acceleration motion. The hybrid WNOJ defined on $SO(3)$ and $\mathbb{R}^{3}$ is more suitable than WNOJ defined on $SE(3)$ to represent this motion. In reality, the motion patterns of objects are more complex than constant acceleration. Furthermore, the trajectory optimization results with these GP assumption are more valuable for our application. To confirm it, we sample the ground truth trajectory of a flying quadrotor with different sampling intervals, and then set an optimization problem with these GP-based priors to infer intermediate states among sampling points. The sampling intervals are set to sampling one ground truth state every $5\sim 160$ points on the trajectory and the trajectory error is calculated with \eqref{gp_traj_error}. As shown in Fig.~\ref{gp_prior_comparison_fpv}, the trajectory errors of WNOA are  significantly greater than WNOJ-local and WNOJ-hybrid. Moreover, as the sampling becomes sparser, the divergence becomes increasingly pronounced.

\subsection{Comparative Analysis of Various Inertial Factors} \label{sec:analy_inertial_factors}

\begin{figure}[!t]
	\centering
	\includegraphics[width=3.2in]{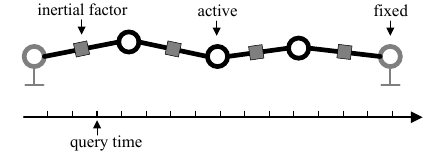}
	\caption{Illustration of inertial factor evaluation. }
	\label{illustrat_inertial_eval}
\end{figure}

From the foregoing discussion, we know that the discrete-preintegration and GPIF rely on the GP-based  trajectory with the WNOJ prior on $SE(3)$ to fuse asynchronous event measurements. 
In contrast, the GPP equivalently induces an inertial-driven motion trajectory based on GP regression, and then uses this trajectory to integrate asynchronous event measurements. We can divide the whole fusion procedure into two steps. 
Firstly, the inertial measurements deform the continuous-time trajectory when they are applied with different types of inertial factors. 
Secondly, the asynchronous feature trajectories are attached on the deformed motion trajectory and continuously refined according to reprojection errors. 
A deformed trajectory that more closely aligns with reality will yield a more precise visual projection at asynchronous measurement times and enable more accurate trajectory estimation thereafter.
Therefore, we evaluate the deformed motion trajectories when applying different inertial factors.

\begin{figure}[!t]
	\centering
	\includegraphics[width=3.3in]{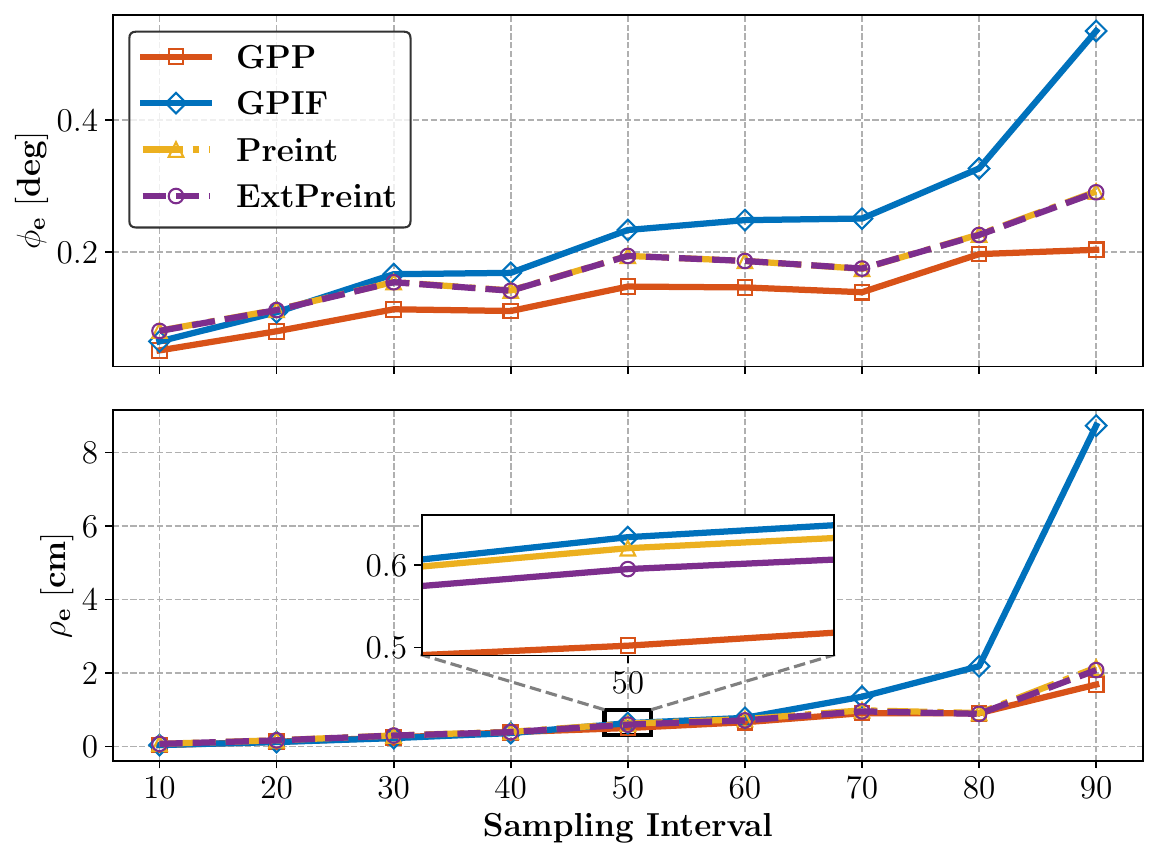}
	\caption{GP trajectory errors of different inertial factors under various sampling intervals. }
	\label{gp_inertial_traj_comparison_fpv}
\end{figure}

We select the \emph{indoor\_45\_9} sequence from UZH-FPV dataset (introduced in Sec.~\ref{sec:high-speed}) and sparsify its ground truth poses by sampling a constant sampling interval (every 10 poses in our evaluation). The original ground truth poses and IMU measurements are recorded at frequencies of approximately  $500\ Hz$ and $1000\ Hz$, respectively.
An inertial-only optimization is performed by further downsampling the sparse poses, keeping them fixed while optimizing the unsampled states.
As shown in Fig.~\ref{illustrat_inertial_eval}, the resulting GP trajectory is queried at the original frequency of the ground truth. We test different downsampling intervals and calculate trajectory errors using \eqref{gp_traj_error}. As displayed in Fig.~\ref{gp_inertial_traj_comparison_fpv}, the GPP achieves the best trajectory precision than others in all tested sampling intervals. The trajectory errors of the GPIF increase rapidly, indicating that the GPIF is more sensitive to variations in sampling intervals. By introducing additional factors for rotational velocities and linear accelerations, the precision of extended preintegration (labeled as ExtPreint) shows slight improvement compared to the preintegration schemes (labeled as Preint). When the sampling interval is quite small (eg: interval $=10$ and $=20$ in Fig.~\ref{gp_inertial_traj_comparison_fpv}), the GPIF achieve lower rotation errors than the Preint and ExtPreint methods. This phenomenon mainly results from the low-pass filtering effect of the GPIF. When the interval is small enough, it suppresses the noise from inertial measurements.

\begin{figure}[!t]
	\centering
	\includegraphics[width=3.4in]{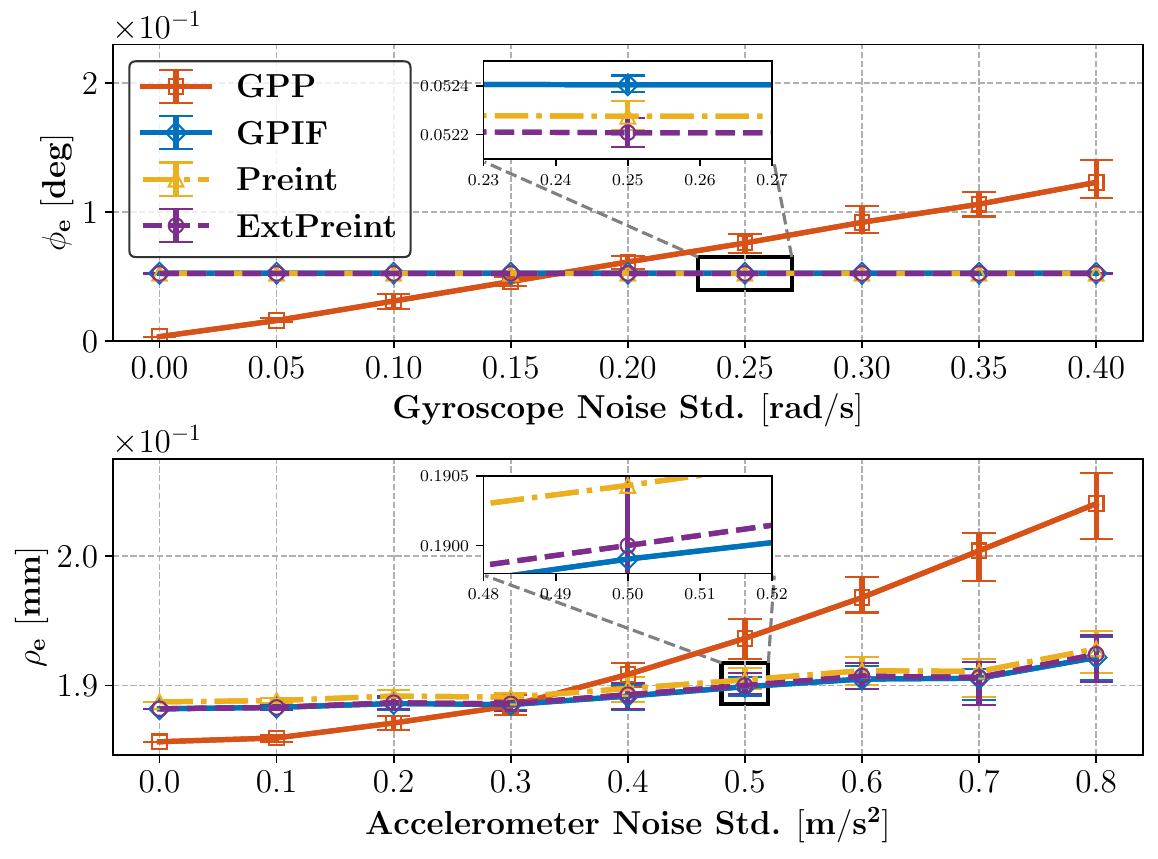}
	\caption{GP trajectory errors of different inertial factors under various Gaussian noises. }
	\label{gp_noisy_inertial_traj_comparison_fpv}
\end{figure}

To further investigate the effect of sensor noises, we disturb the original inertial measurements with zero-mean Gaussian signals. The disturbed inertial measurements are then used to execute the inertial-only optimization (as in Fig.~\ref{illustrat_inertial_eval}). With gradually increasing the covariance of noises,  the resulting trajectory errors are also computed with \eqref{gp_traj_error}. To mitigate the impact of random deviation, each experimental set is repeated $20$ times, with independently noise sampling for each repetition. For a fair comparison, we maintain consistent GP parameters throughout all experiments, and align the IMU parameters in GPP with other methods. From Fig.~\ref{gp_noisy_inertial_traj_comparison_fpv}, the estimated trajectory of GPP starts with the lowest $\phi_{e}$ and $\rho_{e}$ (orange line in Fig.~\ref{gp_noisy_inertial_traj_comparison_fpv}), but degenerates quickly when the covariance of noises becomes larger. Since the GPP induces these latent states in a data-driven manner, its trajectory query precision is highly dependent on the quality of the raw inertial measurements. These results indicate that the GPP may not be the optimal choice for systems equipped with low-frequency, low-quality IMU sensors. In contrast,  the GPIF, Preint, and ExtPreint schemes demonstrate robustness against sensor noises. (three overlapping curves in Fig.~\ref{gp_noisy_inertial_traj_comparison_fpv}). Furthermore, the ExtPreint benefits from the combination of GPIF and Preint, achieving lower rotation and translation errors (as displayed in the magnified regions of Fig.\ref{gp_noisy_inertial_traj_comparison_fpv}). Unlike trajectory query errors, the GPP consistently demonstrates superior accuracy in terms of preintegration compared to discrete preintegration methods.

\subsection{Comprehensive Evaluation in Challenging Scenarios}

\begin{table}[!t]
	\centering
	\renewcommand{\arraystretch}{1.2}
	\caption{Configurations of Evaluated Datasets}
	\label{config_table_datasets}
	\resizebox{3.5in}{!}{
		\begin{threeparttable}
			\begin{tabular}{ l  ccc}
				\toprule[1.5pt]
				Dataset 			&  	Event Camera & Resolution  & Scenario \\
				\midrule
				DAVIS240C\cite{mueggler2017event}			&	DAVIS 240C & 240*180 & Hand-held  \& HDR Indoor\\
				
				UZH-FPV\cite{delmerico2019we}			&	DAVIS 346C & 346*260 & Agressive Indoor \& Outdoor Flying \\
				
				VECtor\cite{gao2022vector} 	&	Prophesee Gen3 & 640*480 & Agressive \& HDR Indoor\\
				
				vicon\_hdr\cite{Guan2023plevio} &	DVXplorer & 640*480 & HDR Indoor\\
				Stereo-HKU\cite{chen2023esvio} &  DAVIS 346C & 346*260 & Agressive \& HDR Indoor\\
				
				Own-collected &	DVXplorer & 640*480 & Indoor \& Outdoor Flying\\
				
				\bottomrule[1.5pt]
			\end{tabular}
			
		\end{threeparttable}
	}
\end{table}

Intuitively, since the AsynEIO discards intensity frames and directly tracks asynchronous event features from event streams, the estimation performance will be benefited from the low motion blur and high HDR nature of event cameras. Therefore, we conduct experiments  on public datasets with similar environments to highlight the performance of AsynEIO. The configurations of these datasets are summarized in Table~\ref{config_table_datasets}. Each of them involves challenging sequences for the monocular event-inertial or visual-inertial odometry. 
Different inertial fusion schemes (as shown in Fig.~\ref{multi_inertial_factor_graph}) are applied to the pipeline of AsynEIO. For accuracy evaluation, we perform alignment of the estimated and ground truth trajectories over the initial $200$ samples ($5\sim 7\ s$ as done in \cite{Guan2023plevio, Vidal2018Ultimate}) of positions and yaw angles. Subsequently, the Absolute Trajectory Error (ATE) is calculated using \eqref{gp_traj_error}. As summarized in Table~\ref{config_methods}, the compared methods include state-of-the-art event-inertial odometry methods based on both traditional (e.g., USLAM, Mono-EIO, and PL-EVIO) and learning-based (e.g., DEVO and DEIO) frameworks. 

\begin{table}[!t]
	\centering
	\renewcommand{\arraystretch}{1.2}
	\caption{Configurations of Different Comparison Methods}
	\label{config_methods}
	\resizebox{3.5in}{!}{
		\begin{threeparttable}
			\begin{tabular}{ l  lcccc}
				\toprule[1.5pt]
				Method 			&  Modality  & Pretreat & Feature & Tracking & Temporal Domain \\
				\midrule
				USLAM\cite{Vidal2018Ultimate} &	E+I & Event Frame & Point & Optical Flow &  Discrete \\
				
				Mono-EIO \cite{guan2022monocular} &	E+I & Time Surface & Point & Optical Flow &  Discrete \\
				
				DEVO \cite{klenk2024deep} & E & Voxel Grid &  Point & Learning-based  & Discrete \\
				
				DEIO \cite{guan2024deio} & E+I & Voxel Grid &  Point & Learning-based & Discrete \\
				
				PL-EVIO \cite{Guan2023plevio} 	&E+F+I & Time Surface & Point \& Line & Optical Flow + LSD &  Discrete \\
				
				Ours &	E+I & Raw & Point & Asyn-front-end  & Continuous \\ 
				\bottomrule[1.5pt]
			\end{tabular}
			
		\end{threeparttable}
	}
\end{table}
\begin{table}[!t]
	\centering
	\renewcommand{\arraystretch}{1.2}
	\caption{The ATE RMSE of DAVIS240C \cite{mueggler2017event}}
	\label{rmse_table_uzh246C}
	\resizebox{3.5in}{!}{
		\begin{threeparttable}
			\begin{tabular}{ l cccccccc}
				\toprule[1.5pt]
				Sequence 			& USLAM & Mono-EIO	& DEIO  		&Preint				& ExtPreint 		&   GPIF  			& GPP          		&	$\text{GPP}^{\ast}$  \\
				\midrule	
				boxes\_6 			& 0.488	& 	0.767	&\textbf{0.053} &0.370 				& \underline{0.364}	& \ding{55}   		& 0.551 	   		&   \ding{55} 	\\
				boxes\_t 	 		& 0.470 & 	0.298	&\textbf{0.118}	&0.203 				& \underline{0.201}	& 0.658   			&0.382		   		&	1.854  	\\
				poster\_6 			& 0.154	& 	0.194	&\textbf{0.050}	&\underline{0.143}	& 0.145				&0.232				&0.278		   		&   0.681   \\
				poster\_t			& 0.211	& 	0.307	&\textbf{0.024}	&\underline{0.101}	&\underline{0.101}	& 0.231   			&0.210		   		&   0.188 \\
				dynamic\_6 			&0.059	& 	0.107	&\textbf{0.022}	&0.068 				& 0.069 			& 0.102  			&\underline{0.048}	& 	0.083  	\\
				dynamic\_t  		&0.471	& 	0.114 	&\textbf{0.024} &0.061 				& 0.061				& 0.122  			&\underline{0.053}	& 	0.094  	\\
				hdr\_boxes	 		& 0.469	& 	0.325	&\textbf{0.067}	&0.235				& \underline{0.126}	& 0.443   			&0.444		   		&   0.951 	\\
				hdr\_poster  		&0.346	& 	0.200 	&\textbf{0.044} &0.184				& 0.186				& 0.151 			&\underline{0.137}	&	0.668 	\\
				shapes\_6	 		& 0.548	& 	/		&  /			&0.515  			& \textbf{0.344}	& 0.928   			&\underline{0.358}	&   \ding{55} 	\\
				shapes\_t			&1.415	& 	/		&	/			&1.344 				& \underline{1.341}	& \ding{55}  		&\textbf{0.184}		& 	\ding{55}	 \\
				\midrule
				Avg. 				& 0.463 &	/		&	/			& 0.322 			& \underline{0.294} & / 				& \textbf{0.265} 	& / \\
				\bottomrule[1.5pt]
			\end{tabular}
			Unit: [m]. The top and second best results are marked in bold and underline, respectively. 
			``\ding{55}'' indicates the method failed to converge and ``/" means the result is not available. 
			The estimated trajectories of USLAM \cite{Vidal2018Ultimate} are generated 
			using publicly available open-source code, while the results for Mono-EIO \cite{guan2022monocular} 
			and DEIO \cite{guan2024deio} are directly obtained from raw estimated trajectories.
		\end{threeparttable}
	}
\end{table}

\subsubsection{Basic Scenarios} The experiments for basic scenarios are conducted on the 
DAVIS240C dataset \cite{mueggler2017event} consisting of event streams from a hand-held DAVIS 240C camera. Among them, \emph{hdr\_boxes} and \emph{hdr\_poster} are captured in low-illumination environments, which can also verify the performance of AsynEIO in HDR environments. Event streams are recorded in the ROS-bag format and republished at $30\ Hz$. The inertial measurements are published with $1000\ Hz$ and the ground truth is captured from Optitrack with $200\ Hz$. 

As shown in Table~\ref{rmse_table_uzh246C}, the proposed framework totally realize a better estimation accuracy than Ultimate-SLAM (labeled as USLAM) \cite{Vidal2018Ultimate} and Mono-EIO \cite{guan2022monocular}. Owing to the online fine-tuning of tracking results using the learning module and parallel computing using GPU, DEIO achieves the highest trajectory accuracy on all sequences of DAVIS240C. 
Generally, GPP and ExtPreint outperform other fusion schemes.
Since the WNOJ assumes a motion trajectory with zero-mean Gaussian jerks, the GPIF can only estimate an approximate mean linear acceleration between adjacent trajectory states. 
This effect significantly reduces the temporal-resolution of raw accelerometer data, limiting it to the sampling frequency of estimated motion states.
Conversely, GPP, Preint, and ExtPreint similarly leverage a preintegration operation between adjacent trajectory states, which largely preserves the original constraints on relative pose and linear velocity. 
Even if the WNOJ assumption may not fully align with reality, these constraints based on the raw observation model help the resulting trajectory match the ground truth. 
By introducing additional constraints on rotational velocity and linear acceleration, ExtPreint enhances the precision of Preint. 
A WNOJ prior attached to the GPP (labeled as $\text{GPP}^{\ast}$ in Table~\ref{rmse_table_uzh246C}) sometimes causes failures. 

\begin{figure*}[thpb]
	\centering
	\includegraphics[width=6.5in]{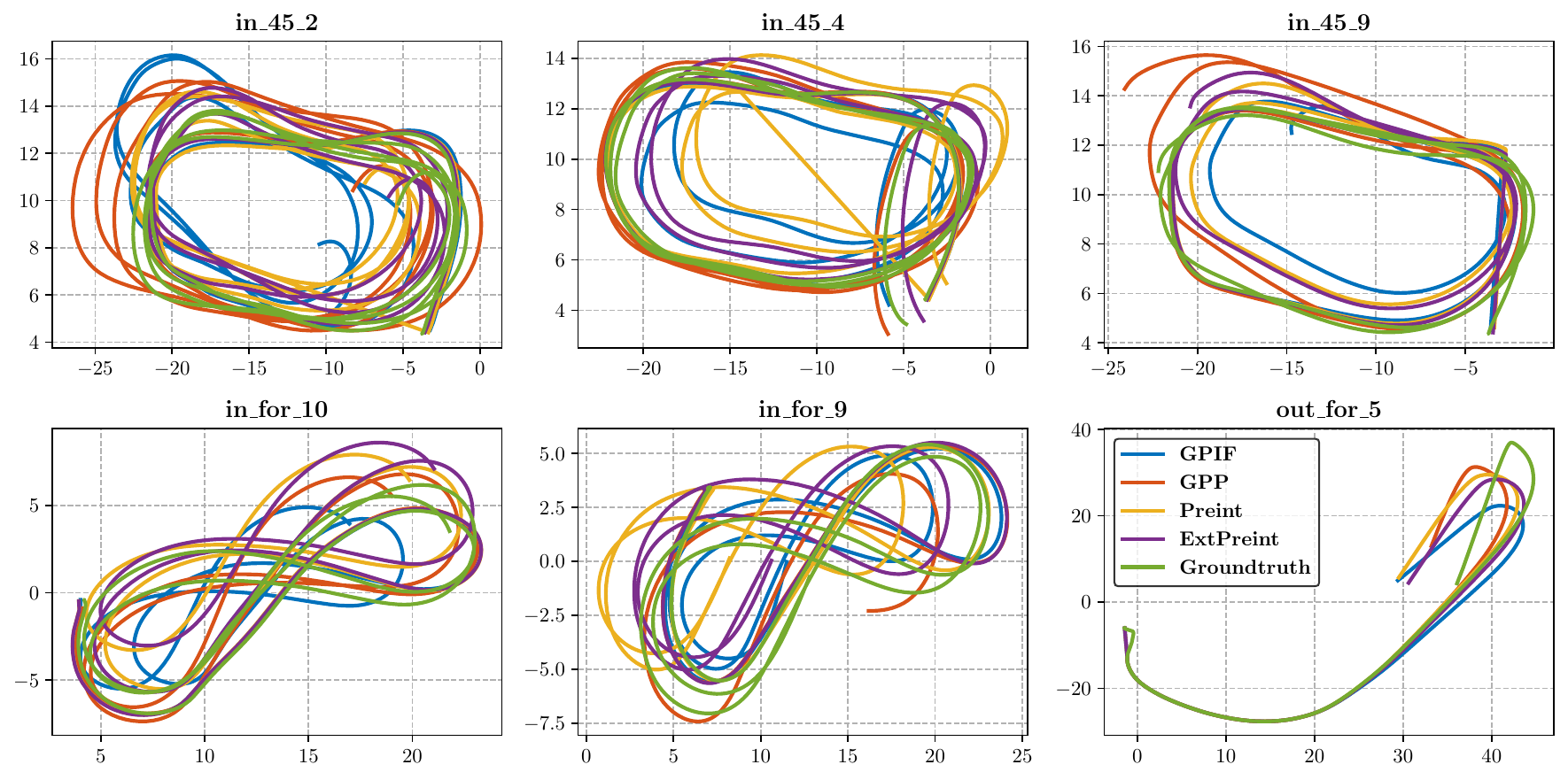}
	\caption{Estimated trajectories for high-speed sequences. }
	\label{overall_trajectory_top_fpv}
\end{figure*}
\begin{table*}[!t]
	\centering
	\renewcommand{\arraystretch}{1.1}
	\caption{The ATE RMSE of High-speed Sequences in UZH-FPV \cite{delmerico2019we}}
	\label{rmse_table_fpv_dataset}
	\resizebox{7.0in}{!}{
		\begin{threeparttable}
			\begin{tabular}{ l cccccccccccccccccc}
				\toprule[1.5pt]
				\multirow{2}{*}{Sequence} & \multicolumn{2}{c}{Velocity} & \multicolumn{2}{c}{USLAM\cite{Vidal2018Ultimate}} & \multicolumn{2}{c}{DEVO\cite{klenk2024deep}} & \multicolumn{2}{c}{DEIO\cite{guan2024deio}} & \multicolumn{2}{c}{Preint} 	& \multicolumn{2}{c}{ExtPreint} 	&   \multicolumn{2}{c}{GPIF}  & \multicolumn{2}{c}{GPP} & \multicolumn{2}{c}{$\text{GPP}^{\ast}$} 		  \\
				\cmidrule(r){2-3} \cmidrule(r){4-5}	\cmidrule(r){6-7}	\cmidrule(r){8-9} \cmidrule(r){10-11} \cmidrule(r){12-13} \cmidrule(r){14-15}  \cmidrule(r){16-17}  \cmidrule(r){18-19}
				& $\nu_{max}$ & $\omega_{max}$ & $\phi_{e}$ & $\rho_{e}$ & $\phi_{e}$ & $\rho_{e}$ & $\phi_{e}$ & $\rho_{e}$ & $\phi_{e}$ & $\rho_{e}$ & $\phi_{e}$ & $\rho_{e}$ & $\phi_{e}$ & $\rho_{e}$ & $\phi_{e}$ & $\rho_{e}$ & $\phi_{e}$ & $\rho_{e}$  \\
				
				\cmidrule(r){1-3}	\cmidrule(r){4-5} \cmidrule(r){6-7} \cmidrule(r){8-9} \cmidrule(r){10-11} \cmidrule(r){12-13} \cmidrule(r){14-15}  \cmidrule(r){16-17} \cmidrule(r){18-19}
				indoor\_45\_2 		&6.97 & 2.55& \ding{55} &\ding{55} & 0.094 & \underline{2.185} & / & / &	\underline{0.066}&3.056 	& 0.082	& \textbf{2.139}&0.315& 4.870   &\textbf{0.040}& 2.542 	& 0.125  & 2.209 	   \\
				indoor\_45\_4 	 	&6.55 & 1.58 & 0.203 & 15.958 & 0.042 & \textbf{0.686} & /& / &	\underline{0.029}&2.394 	&0.354	& 1.366 		&0.051& 1.596   &\textbf{0.019}& \underline{1.034} 	& 0.035  & 1.304  \\
				indoor\_45\_9 		&11.23& 2.81& 0.172 & 5.593 &0.691 & 1.893 &/ & /&	\underline{0.063}&1.541		&0.070	& 1.738			&0.083& 2.379   &0.120& \underline{1.377}  & \textbf{0.061} & \textbf{0.743}   \\
				indoor\_45\_12   	&  4.10  &1.18 & \ding{55}&\ding{55}     &     0.055     &  1.048        &/ &/ &    \textbf{0.019}&\underline{0.622}     &\underline{0.020}  & \textbf{0.607}         &0.102& 1.096   &0.030& 0.667   &0.038   & 0.806 \\
				indoor\_45\_13		&	7.72  &  2.06 & \ding{55}&\ding{55}   &     0.151      &    2.475      &/&/&    \textbf{0.024}&\textbf{0.651}     &\textbf{0.024}  & \underline{0.653}         &\underline{0.038}& 2.097   &0.049&0.988   &0.049   & 1.028 \\
				indoor\_45\_14	&	11.71  & 3.86 & \ding{55}&\ding{55}	    &   \ding{55}   &  \ding{55}     &/ &/&   \textbf{0.045}& \underline{2.557}		&\textbf{0.045}	& \textbf{2.551}			&0.119& 4.411	&\underline{0.081}&2.854   &\underline{0.081}	& 2.840 \\
				indoor\_for\_3	&	9.29  &4.02	&\ding{55}&\ding{55}	 &	\underline{0.106}		&	\textbf{0.907}	   & \textbf{0.083} & 2.770&   0.167&3.591		&0.173  &  \underline{2.515}			&0.254& 3.428  &0.149&3.548   & 0.137	& 4.248 \\
				indoor\_for\_5  &	4.69  &	1.26  &\ding{55}&\ding{55}   &		\underline{0.039}	&	0.487	   & \textbf{0.032} & 0.407 &	0.050&\textbf{0.372}		&0.050	& \underline{0.378}			&0.058& 0.651	& 0.080 &0.836   &0.081	& 0.913 \\
				indoor\_for\_6  &	11.76  & 5.27	&\ding{55}&\ding{55} &	0.148		&	\textbf{0.740}	   &0.128&\underline{2.231}&	\textbf{0.093}&3.233		&\textbf{0.093}	& 3.238			&\underline{0.113}& 3.371	&0.343& 3.463 &0.197	& 2.885 \\
				indoor\_for\_7  &	12.10  &	6.58 &\ding{55}&\ding{55}	 &	0.279 		&	\underline{2.686}	   &\underline{0.234}&3.537&	0.311&3.240	&0.380	& 3.206			&\textbf{0.160}& \textbf{2.286}	& 0.235 & 2.824 &0.687	& 6.991 \\
				indoor\_for\_9  &11.42& 2.08 &\ding{55}&\ding{55} & \textbf{0.034} & \textbf{0.564} &0.056&1.343& \underline{0.054}&3.010&0.059	& 1.911			&0.121& 1.987	&0.070 & 1.531 & 0.062 	& \underline{1.295}   \\
				indoor\_for\_10 & 9.49& 2.11 & \ding{55}&\ding{55} & \underline{0.035} & \textbf{0.756} & 0.037 & \underline{0.994} &   0.041&1.531  	& 0.038	& 1.598			&0.099& 2.114   &0.116& 1.282 &\textbf{0.028}&2.111  \\
				outdoor\_for\_1	&	8.40  &	1.59 &\ding{55}&\ding{55}	 &	0.095	&	\underline{5.000}	& / &  / &	0.053&5.236		&0.053	& \textbf{4.812}			&0.054& 9.812	&\textbf{0.040}& 5.308  &\underline{0.045}	& 9.144 \\
				outdoor\_for\_3 &	11.29  & 2.09 &\ding{55}&\ding{55}  &    0.341      &    8.806     &/& /& 	0.069&\textbf{3.636}		&0.069	& \underline{3.645}			&\ding{55}&\ding{55}	&\underline{0.065}& 5.029  &\textbf{0.055}	& 8.607 \\
				outdoor\_for\_5 &20.73& 1.98 &\ding{55}&\ding{55}& 0.084 & \textbf{1.480} & / & / & \textbf{0.039}&6.050		& \underline{0.044}	& 6.376		&\ding{55}&\ding{55}&0.048& \underline{4.435}       & 0.046 & 5.950      \\
				outdoor\_45\_1		&	11.88  &	2.22 &\ding{55}&\ding{55}	 &	\textbf{0.063}		&	\textbf{2.406}	 & / & / &	0.216& 3.259	& 0.217& 3.270 		&\underline{0.157}& 3.394	&0.191	&\underline{3.198}	&0.369	 & 5.377 \\
				\cmidrule(r){1-3}	\cmidrule(r){4-5} \cmidrule(r){6-7} \cmidrule(r){8-9} \cmidrule(r){10-11} \cmidrule(r){12-13} \cmidrule(r){14-15} \cmidrule(r){16-17}  \cmidrule(r){18-19}
				Avg.    & / & / & / & / & / & / & /& /& \textbf{0.084} & 2.749& 0.111 & \textbf{2.500} & \ding{55}   &  \ding{55} & \underline{0.105} & \underline{2.557} & 0.131 & 3.528  \\
				\bottomrule[1.5pt]
			\end{tabular}
			$\nu_{max}$ is the maximum linear velocity (m/s), $\omega_{max}$ the maximum rotational velocity (rad/s), 
			$\rho_{e}$ the translation error (m), and $\phi_{e}$ the rotation error (rad). The results of USLAM and DEVO are generated using 
			their open-source codes and the results of DEIO are directly obtained from raw estimated trajectories. As the DEVO is a monocular event-based
			method (without IMUs), their estimated trajectories suffer from scale ambiguity and are aligned with the ground truth trajectories in $SIM(3)$.
		\end{threeparttable}
	}
\end{table*} 

\subsubsection{High-speed and Aggressive Maneuvering} \label{sec:high-speed}
The experiments for high-speed scenarios are conducted on three datasets, including UZH-FPV \cite{delmerico2019we} and VECtor \cite{gao2022vector}, and Stereo-HKU \cite{chen2023esvio}.  In the UZH-FPV dataset, event streams are recorded using an mDAVIS camera mounted on a First-Person-View (FPV) racing quadrotor. The maximum linear velocities of the selected sequences range from $4.10\ m/s$ to $20.73\ m/s$. The inertial measurements and event array messages are published at $1000 Hz$ and $30 Hz$, respectively. This dataset comprises high-speed laps around a racetrack featuring drone racing gates, as well as free-form trajectories navigating obstacles in both indoor and outdoor environments. The ground truth is obtained from a Leica Nova MS60 laser tracker in $500\ Hz$.  For the VECtor dataset, Event streams come from two Prophesee Gen3 event cameras, and the ground truth is collected using the Optitrack motion capture system in 120 Hz. As shown in Table~\ref{rmse_table_fpv_dataset} and \ref{rmse_table_vector_dataset}, the VECtor dataset have higher rotational velocities than the UZH-FPV while the UZH-FPV having super high-speed linear velocities. 
The Stereo-HKU dataset comprises sequences collected from a handheld stereo event camera under extremely aggressive motion and strong HDR environments. Their event streams and IMU messages are published in $60\ Hz$ and $1000\ Hz$, respectively. The ground truth comes from a motion capture system at $50\ Hz$.

Table~\ref{rmse_table_fpv_dataset} compares the performance of AsynEIO, USLAM, DEVO, and DEIO on UZH-FPV dataset. Estimated trajectories of AsynEIO are also visualized in Fig.~\ref{overall_trajectory_top_fpv}. USLAM demonstrates the highest estimation errors and fails to operate successfully on most high-speed sequences. In contrast, the other methods exhibit superior robustness in high-speed scenarios.  
As a monocular event odometry method, DEVO exhibits significant scale drifts on \emph{outdoor\_for\_1} and \emph{outdoor\_for\_3}, and fails to estimate on \emph{indoor\_45\_14}. There are only 6 available trajectories for DEIO (indoor forward flying, labeled as indoor\_for\_n in Table~\ref{rmse_table_fpv_dataset}). It is noteworthy that $SIM(3)$ alignment is adopted to eliminate the scale drift in DEVO, while the other methods are aligned using only position and yaw angles. After this alignment, DEVO even outperforms DEIO in terms of accuracy. Compared with DEIO, our AsynEIO achieves higher position precision on \emph{indoor\_for\_3}, \emph{indoor\_for\_5}, \emph{indoor\_for\_7}, and \emph{indoor\_for\_9}. In terms of the attitude estimation metric $\phi_{e}$, AsynEIO consistently outperforms the other methods.
Benefiting from the event-driven front-end, the high temporal resolution of the event camera is well-preserved, enabling effectively feature tracking for high-speed maneuvers.
Meanwhile, directly tracking on raw event streams can partially overcome the motion blur caused by high-speed maneuvers. For instance, our AsynEIO is able to maintain estimation at a super high speed of $20.73\ m/s$ in \emph{outdoor\_for\_5}. The increased position errors are primarily caused by degeneracy in large-scale outdoor environments and by high-frequency texture regions, such as grass surfaces. In longer sequences, such as \emph{indoor\_for\_3} (about $275\ m$) and \emph{outdoor\_for\_3} (about $615\ m$), AsynEIO also suffers from accumulative drifts in the absence of any loop-closure. 
Among the different inertial schemes implemented in AsynEIO, GPP and ExtPreint yield higher positional accuracy, whereas Preint exhibits superior performance in attitude estimation.
Table~\ref{rmse_table_vector_dataset} also summarizes some ATE results of aggressive sequences on VECtor dataset. 
On these sequences, PL-EVIO achieves the best positional precision, which mainly benefits from the use of redundant feature modalities (points and lines), multiple sensor observations (event, frame, and IMU), as well as the presence of structured environments. 
Consistent with the results on UZH-FPV, AsynEIO outperforms both USLAM and PL-EVIO in attitude estimation and significantly improves the position accuracy compared to USLAM.
From the aggressive sequences in Table~\ref{rmse_table_stereo_hku_dataset}, we observe that our AsynEIO achieves the best estimation performance on aggressive rotation data (i.e., \emph{hku\_agg\_rotation}) and yields lowest attitude errors across all sequences. However, DEIO and PL-EVIO achieve better position estimation. Unlike the UZH-FPV dataset, the performance of GPIF becomes less distinct from that of GPP and ExtPreint on the VECtor and Stereo-HKU datasets. This might be because aggressive maneuvering provides more substantial IMU excitation, thereby allowing the residuals of GPIF to play a more effective role in the optimization process.

\begin{figure}[!t]
	\centering
	\includegraphics[width=2.8in]{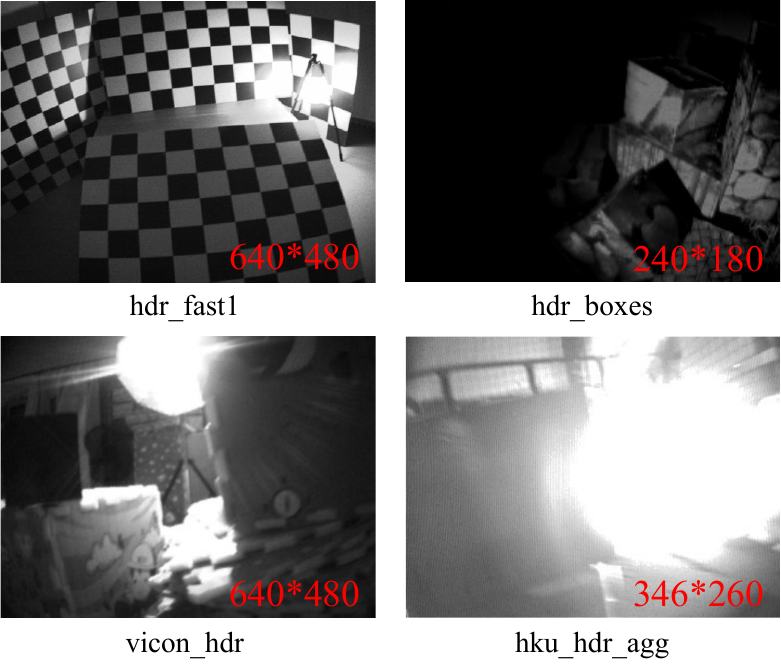}
	\caption{Sequences in challenging illumination scenarios.}
	\label{hdr_images_four_seqs}
\end{figure}

\begin{figure}[!t]
	\centering
	\includegraphics[width=3.4in]{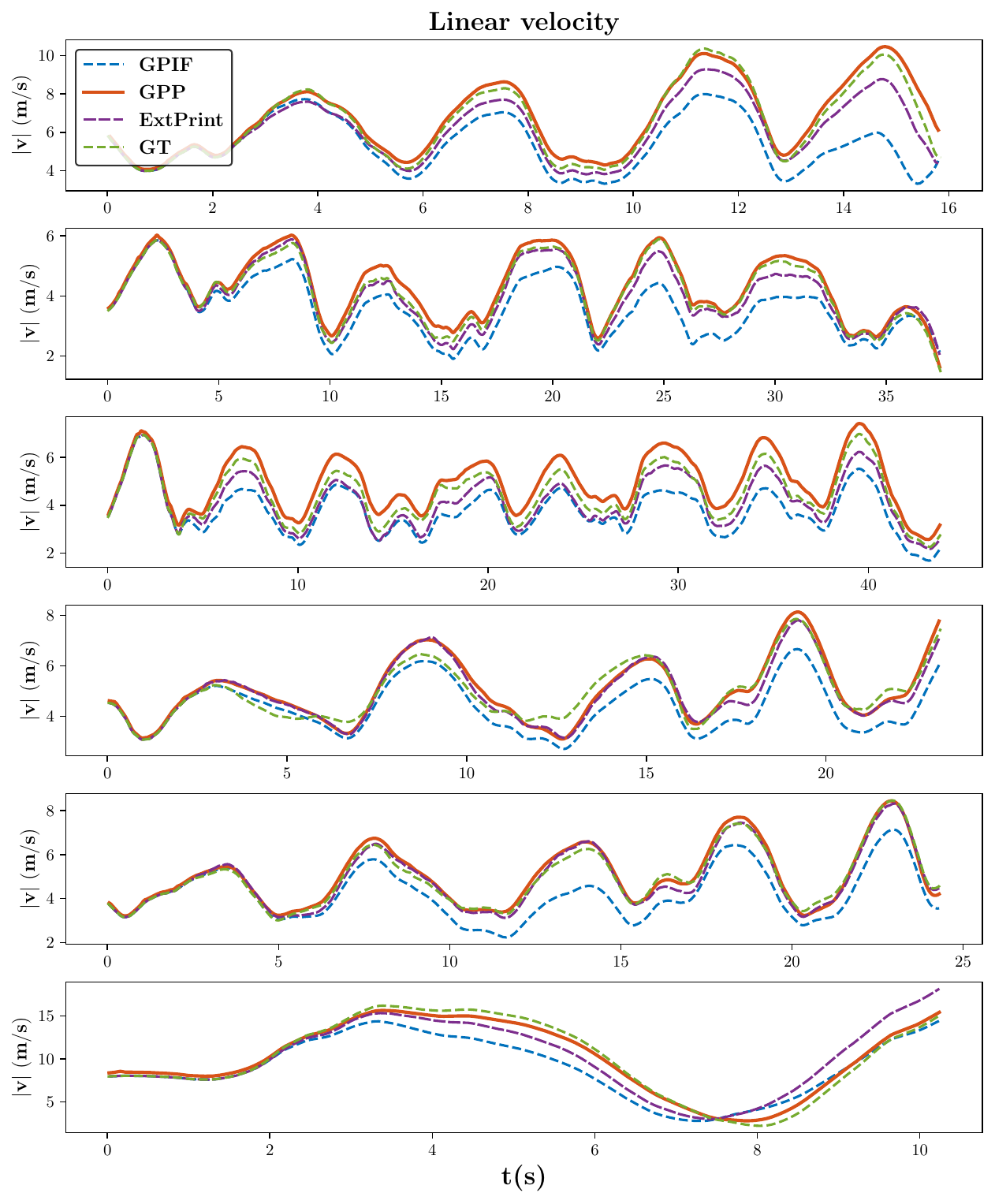}
	\caption{Estimated linear velocity on FPV dataset, including \emph{indoor\_45\_2}, \emph{indoor\_45\_4}, \emph{indoor\_45\_9}, \emph{indoor\_for\_9}, \emph{indoor\_for\_10} and \emph{outdoor\_for\_5} (from top to bottom).}
	\label{estimated_velocity}
\end{figure}

\subsubsection{Challenging Illumination} \label{sec:hdr_eval}
The performance in challenging illumination environments can be validated
through HDR sequences from VECtor \cite{gao2022vector}, Stereo-HKU \cite{chen2023esvio}, and \emph{vicon\_hdr} \cite{Guan2023plevio} (as listed in Table~\ref{config_table_datasets}).  
These HDR scenarios are shown in Fig.~\ref{hdr_images_four_seqs} to intuitively illustrate these challenging illumination conditions. 
The resulting ATE can be found in Table~\ref{rmse_table_vector_dataset} and Table~\ref{rmse_table_stereo_hku_dataset}.
The learning-based DEIO and the multi-modality PL-EVIO surpass our AsynEIO in terms of the $\rho_{e}$ metric on HDR sequences in the Stereo-HKU dataset. However, our pipeline, when using ExtPreint, Preint, or GPIF, outperforms the others on the \emph{vicon\_hdr} sequence.  
In addition, AsynEIO still has apparent advantages in handling rotational error $\phi_{e}$, such as \emph{hku\_hdr\_tran\_rotate}, \emph{hku\_hdr\_agg} and \emph{hku\_dark\_normal} in Table.~\ref{rmse_table_stereo_hku_dataset}. For high dynamic range scenarios, our event-based front-end demonstrates robustness to stable low-light and high-light regions (e.g., \emph{hdr\_boxes} and \emph{vicon\_hdr} in Fig.~\ref{hdr_images_four_seqs}). However, it is less resilient to sudden, global illumination changes in the environment, which can cause a temporary ``blinding'' effect (e.g., \emph{hku\_hdr\_agg} in Fig.~\ref{hdr_images_four_seqs}).

\subsubsection{Velocity Estimation} 

Fig.~\ref{estimated_velocity} presents the estimated velocities for the sequences \emph{indoor\_45\_2}, \emph{indoor\_45\_4}, \emph{indoor\_45\_9}, \emph{indoor\_for\_9}, \emph{indoor\_for\_10}, and \emph{outdoor\_for\_5} from the UZH-FPV dataset.
Overall, the velocities derived from the GPP method (orange solid curves) align more closely with the ground truth (green dashed curves), whereas the GPIF method (blue dashed curves) generally underestimates the magnitude of the velocities. 
In most sequences, In most sequences, ExtPreint yields results comparable to those of GPP. 
Accurate and high-frequency velocity estimation is crucial for the feedback controllers of robots.
In practice, discrete-time VIO systems typically operate at a limited update frequency (10 Hz to 30 Hz).  
To meet the requirements of feedback control (normally $\ge 100\ Hz$ for a drone), researchers often predict high-frequency states by integrating the latest state estimations with inertial measurements.
Until the next state is estimated, the velocity and pose are adjusted and may ``jump'' to the new estimation.
This process can lead to significant inconsistencies due to the integration of noisy inertial measurements.
In contrast, our asynchronous estimation framework enables continuous inference of high-frequency states through a tightly-coupled fusion of event and inertial measurements. 
Consequently, our proposed method holds great potential for enhancing state feedback in controllers, thereby improving control for rapidly maneuvering robots.

\begin{table*}[!t]
	\centering
	\renewcommand{\arraystretch}{1.2}
	\caption{The ATE RMSE of Aggressive and HDR Sequences in \cite{gao2022vector, Guan2023plevio} }
	\label{rmse_table_vector_dataset}
	\resizebox{7.0in}{!}{
		\begin{threeparttable}
			\begin{tabular}{ l cccccccccccccccc}
				\toprule[1.5pt]
				\multirow{2}{*}{Sequence} 	& \multicolumn{2}{c}{Velocity} 	& \multicolumn{2}{c}{USLAM \cite{Vidal2018Ultimate}} 	& \multicolumn{2}{c}{PL-EVIO\cite{Guan2023plevio}} 	& \multicolumn{2}{c}{Preint} 		& \multicolumn{2}{c}{ExtPreint} 		&   \multicolumn{2}{c}{GPIF}  		& \multicolumn{2}{c}{GPP} 			& \multicolumn{2}{c}{$\text{GPP}^{\ast}$} 		  \\
				\cmidrule(r){2-3} 				\cmidrule(r){4-5}										\cmidrule(r){6-7}									\cmidrule(r){8-9} 					\cmidrule(r){10-11} 					\cmidrule(r){12-13} 				\cmidrule(r){14-15} 				\cmidrule(r){16-17} 
				& $\nu_{max}$ & $\omega_{max}$	& $\phi_{e}$ & $\rho_{e}$ 								& $\phi_{e}$ & $\rho_{e}$ 							& $\phi_{e}$ & $\rho_{e}$ 			& $\phi_{e}$ & $\rho_{e}$ 				& $\phi_{e}$ & $\rho_{e}$ 			& $\phi_{e}$ & $\rho_{e}$ 			& $\phi_{e}$ & $\rho_{e}$  \\
				
				\cmidrule(r){1-3}											\cmidrule(r){4-5} 										\cmidrule(r){6-7} 									\cmidrule(r){8-9} 					\cmidrule(r){10-11} 					\cmidrule(r){12-13} 				\cmidrule(r){14-15}  				\cmidrule(r){16-17}
				desk\_fast 				&1.44 & 3.98					&  0.492	&0.914	  									& 0.271 & \textbf{0.108}  							&  0.380	& 1.268 				& \underline{0.162}	& 0.689 			&0.307&  1.156  					&0.250&  0.566						& \textbf{0.151} & \underline{0.480}   \\
				robot\_fast  				&1.52 & 4.42					&  0.119	&0.944	  									& 0.170 & \textbf{0.045} 							&  0.079	& 0.481 				&0.079	& 0.484							&0.311&  0.421  					&\textbf{0.067} & \underline{0.190} & \underline{0.077} & 0.396   \\
				mountain\_fast 			&1.26 & 3.46					&  0.262	&0.565	  									& 0.149 & \textbf{0.060}  							&  0.097	& 0.150 				& 0.121	& \underline{0.135}				&0.118&  0.191  					& \underline{0.086} & 0.247 		& \textbf{0.071} & 0.373  \\
				sofa\_fast   				& 1.73 & 4.51					&  0.427	& 0.602	  								& 0.208 & \textbf{0.085} 								&  0.188	& 0.470 				& 0.190	& 0.478 						&\textbf{0.074} &  \underline{0.367}  					& 0.198 & 0.450						& \underline{0.185} & 0.455  \\
				\cmidrule(r){1-3}			\cmidrule(r){4-5} 				\cmidrule(r){6-7} 										\cmidrule(r){8-9} 									\cmidrule(r){10-11} 				\cmidrule(r){12-13} 					\cmidrule(r){14-15} 				\cmidrule(r){16-17} 
				hdr\_fast 					& 0.91	& 2.11						&0.662		&  0.699 									& 0.132 & \textbf{0.068} 							&0.029& 0.252 			& \underline{0.024} &\underline{0.223} 	&0.153&  0.383 						&0.344& 0.404 						& \textbf{0.021} & \underline{0.223}	 \\
				vicon\_hdr 					& 1.51 & 5.04							&0.212		&  2.235									& 0.149 & 0.584 						&\textbf{0.043}& \underline{0.361}  &0.111&  \textbf{0.354} 				&\underline{0.092}&  0.374  		&0.290& 0.746  						& \ding{55} &\ding{55} 	 \\
				\cmidrule(r){1-3}			\cmidrule(r){4-5} 				\cmidrule(r){6-7} 										\cmidrule(r){8-9} 									\cmidrule(r){10-11} 				\cmidrule(r){12-13} 					\cmidrule(r){14-15} 				\cmidrule(r){16-17} 
				Avg.   				 		& / & / 						& 0.363 & 0.993 										& 0.180 & \textbf{0.158} 									& \underline{0.136} & 0.497						& \textbf{0.115} & \underline{0.394} 			&0.176 & 0.482 						& 0.206 & 0.434 & / & /  \\
				\bottomrule[1.5pt]
			\end{tabular}
			$\nu_{max}$ is the maximum linear velocity (m/s), 
			$\omega_{max}$ the maximum rotational velocity (rad/s), 
			$\rho_{e}$ the translation error (m), and $\phi_{e}$ the rotation error (rad). 
			The baseline trajectories of PL-EVIO are taken from \cite{Guan2023plevio} and the results of USLAM are 
			generated using the public available open-source code. The PL-EVIO utilizes a combination of events and frames where both 
			point and line features are tracked and fused with inertial measurements. 
			
		\end{threeparttable}
	}
\end{table*} 

\begin{table*}[!t]
	\centering
	\renewcommand{\arraystretch}{1.2}
	\caption{The ATE RMSE of Aggressive and HDR Sequences in Stereo-HKU \cite{chen2023esvio}  }
	\label{rmse_table_stereo_hku_dataset}
			\resizebox{7.0in}{!}{
		\begin{threeparttable}
			\begin{tabular}{ l   cc  cc  cc  cc cc  cc  cc  cc}
				\toprule[1.5pt]
				\multirow{2}{*}{Sequence}  &  \multicolumn{2}{c}{USLAM \cite{Vidal2018Ultimate}}	&\multicolumn{2}{c}{DEIO\cite{guan2024deio}}	& \multicolumn{2}{c}{PL-EVIO \cite{Guan2023plevio}} 	& \multicolumn{2}{c}{Preint} 			& \multicolumn{2}{c}{ExtPreint}		& \multicolumn{2}{c}{GPIF}	& \multicolumn{2}{c}{GPP} 	& \multicolumn{2}{c}{$\text{GPP}^{\ast}$}  \\
				\cmidrule(r){2-3}									\cmidrule(r){4-5} 									\cmidrule(r){6-7} 						\cmidrule(r){8-9} 					\cmidrule(r){10-11} 		\cmidrule(r){12-13} 		\cmidrule(r){14-15}  \cmidrule(r){16-17}
				& $\phi_{e}$& $\rho_{e}$  & $\phi_{e}$& $\rho_{e}$ 							& $\phi_{e}$ & $\rho_{e}$ 							& $\phi_{e}$ & $\rho_{e}$ 				& $\phi_{e}$ & $\rho_{e}$ 			& $\phi_{e}$ & $\rho_{e}$ 	& $\phi_{e}$ & $\rho_{e}$ 	& $\phi_{e}$ & $\rho_{e}$ \\
				\cmidrule(r){1-3}																\cmidrule(r){4-5} 									\cmidrule(r){6-7} 						\cmidrule(r){8-9} 					\cmidrule(r){10-11} 		\cmidrule(r){12-13}  		\cmidrule(r){14-15}   \cmidrule(r){16-17}
				hku\_agg\_translation		& 0.499 & 10.304  & 0.064 & \underline{0.091} 						&0.136 & \textbf{0.074}								&0.042 & 0.211		 					&0.043 & 0.210						&0.049 & 0.240 				& \textbf{0.033} & 0.177 	& \underline{0.038} & 0.211 \\
				hku\_agg\_rotation		& \ding{55}  & \ding{55}	& 0.089 & \underline{0.138} 						&0.216 & 0.229										&0.101 & 0.202		 					&0.101 & 0.201						&0.061 & 0.235 				& \underline{0.048} & 0.257	& \textbf{0.040} & \textbf{0.127} \\
				hku\_agg\_flip		& 1.371 &	4.150	& 0.185 & \textbf{0.222} 							&0.407 & \underline{0.243}							&\textbf{0.161} & 0.948		 			&\textbf{0.161} & 0.948				&0.171 & 1.170 				& 0.172 & 0.796				& \underline{0.166} & 1.701 \\
				hku\_agg\_walk		&   \ding{55}  &	\ding{55} 	& 0.158 & \underline{0.683} 						&0.205 & \textbf{0.658}								&0.043 & 2.963		 					&\underline{0.042} & 2.977			&0.087 & 1.069 				& \textbf{0.033} & 4.596 	& 0.093 & 4.081 \\
				\cmidrule(r){1-3}																\cmidrule(r){4-5} 									\cmidrule(r){6-7}						\cmidrule(r){8-9} 					\cmidrule(r){10-11} 	 	\cmidrule(r){12-13} 		\cmidrule(r){14-15}   \cmidrule(r){16-17}
				hku\_hdr\_circle	 &  0.495 &	0.450	& \underline{0.119}	& \underline{0.195}				& 0.201 & \textbf{0.091}							&0.128 & 0.458							&0.128 & 0.458		 				&0.154 & 0.958				&\textbf{0.099} & 0.841 	& \textbf{0.099} & 1.318  \\
				hku\_hdr\_slow		&  \ding{55} &	\ding{55}	& \textbf{0.064}	& \textbf{0.078}				& 0.086	& \underline{0.114}							&0.482 & 1.926							&0.483 & 1.924 		 				&0.330 & 1.632				&\underline{0.076} & 2.951  & 0.293 & 1.948  \\
				hku\_hdr\_tran\_rotate	& \ding{55} & \ding{55}	& \underline{0.062}	& \underline{0.141}				& 0.097	& \textbf{0.091}							&\textbf{0.042} & 0.276					&\textbf{0.042} & 0.277		 		&0.105 & 0.565				&0.159& 0.644 				& 0.137 & 0.578  \\
				hku\_hdr\_agg		&  \ding{55} &	 \ding{55}	& \underline{0.105}	& \textbf{0.132}				& 0.248 & \underline{0.447}							&0.163 & 4.680							&0.152 & 4.684		 				& \textbf{0.078} &1.739		&0.273& 0.942 				& 0.313 & 4.610  \\
				hku\_dark\_normal	& \ding{55}  &	\ding{55}	& \underline{0.059}	& \textbf{0.187}				& 0.249 & 2.534										&0.079 & 0.562							&0.079 & 0.563		 				&\textbf{0.038}&\underline{0.520}&0.301& 0.893 			& 0.309 & 0.883  \\
				\cmidrule(r){1-3}																\cmidrule(r){4-5} 									\cmidrule(r){6-7}						\cmidrule(r){8-9} 					\cmidrule(r){10-11} 	 	\cmidrule(r){12-13} 		\cmidrule(r){14-15}   \cmidrule(r){16-17}
				Avg. 			&  / &	/	& \textbf{0.101}	& \textbf{0.207} 				&0.205 	& \underline{0.498}							&0.138 & 1.296		 					&0.137 & 1.360						&\underline{0.119} & 0.903 				& 0.132 & 1.344 			& 0.165 & 1.717 \\
				\bottomrule[1.5pt]
			\end{tabular}
			$\rho_{e}$ and $\phi_{e}$ are the translation (m) and rotation (rad) error computed 
			using \eqref{gp_traj_error}. The results of USLAM are 
			generated using the public available open-source code. The estimated trajectories of DEIO and PL-EVIO are taken from \cite{guan2024deio} and \cite{Guan2023plevio} respectively.
		\end{threeparttable}
				}
\end{table*}

\subsection{Own Collected Date Results}

\begin{figure}[!t]
	\centering
	\includegraphics[width=3.0in]{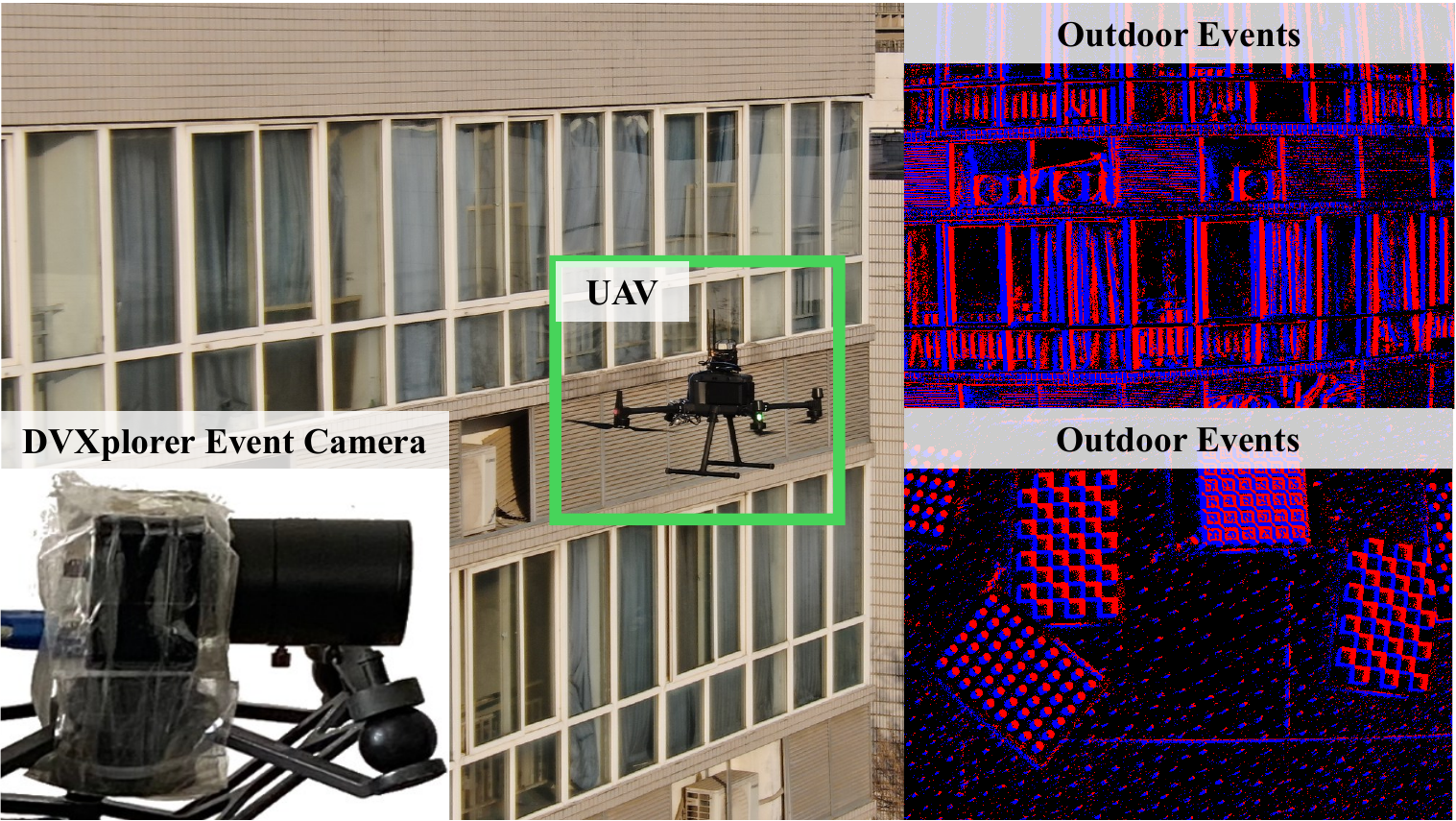}
	\caption{Experimental equipment and  scenarios for own-collected data.}
	\label{uav_dvxplorer}
\end{figure}

To further validate the proposal in realistic environments, we collect event-inertial sequences using a quadrotor (DJI M300-RTK) equipped with 
a DVXplorer ($640 \times 480$ pixels) event camera (as illustrated in Fig.~\ref{uav_dvxplorer}).
The inertial measurements are gathered from the quadrotor at $400\ Hz$ and from the event camera at $1000\ Hz$, while the event streams are collected at $30\ Hz$. 
The IMU of the quadrotor has non-strict software synchronization, while the IMU of the event camera features strict hardware-level event synchronization.
This own-collected sequences consists of two indoor 
and five outdoor sequences. In indoor scenarios, the quadrotor is manually maneuvered within 
a motion-capture room for event-inertial data collection, with ground truth provided at $50\ Hz$. 
In outdoor scenarios, the quadrotor is controlled to fly towards buildings (as displayed in Fig.~\ref{uav_dvxplorer}), with ground truth data 
recorded using a high-precision RTK system. Since the RTK can only offer a low-frequency ($5\ Hz$), we fuse it with a visual-inertial odometry \cite{wang2024localization} to realize stable high-frequency ground truth in $50\ Hz$.
The onboard computer equipped with an Intel Core i7-8550U CPU @ 1.80 GHz.  Intrinsic and extrinsic parameters of the DVXplorer and drone are calibrated using E-calib \cite{SalahEcalib} and Kalibr\footnote{https://github.com/ethz-asl/kalibr}. 

\begin{figure}[!t]
	\centering
	\includegraphics[width=3.2in]{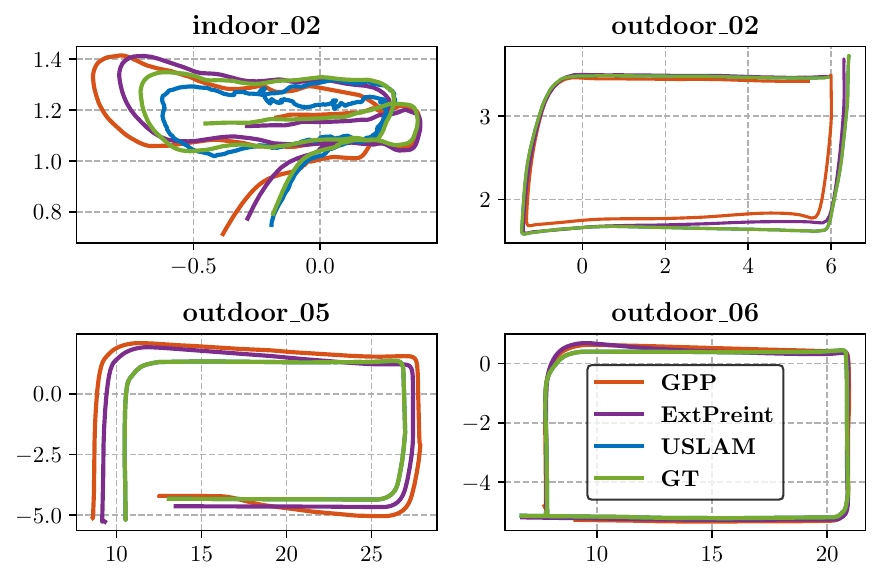}
	\caption{Estimated trajectory on our collecting sequences.}
	\label{overall_trajectory_top_own_collected}
\end{figure}

\begin{table}[!t]
	\centering
	\renewcommand{\arraystretch}{1.2}
	\caption{The ATE RMSE of Own-Collected Sequences}
	\label{The RMSE of Own Collected dataset}
	\resizebox{3.5in}{!}{
		\begin{threeparttable}
			\begin{tabular}{ l c  c  c  c  c  c }
				\toprule[1.5pt]
				
				Sequence 		& USLAM \cite{Vidal2018Ultimate}	& Preint 	& ExtPreint	& GPIF	  & GPP 	& $\text{GPP}^{\ast}$  	 \\
				
				& \ ev 	/  uav 					& \ ev 	/  uav 			& \ ev 	/  uav 			& \ ev 	/  uav 			& \ ev 	/  uav 				& \ ev 	/  uav 	\\
				\midrule
				indoor\_02 		&  0.286 / 0.404				& 0.106 / \textbf{0.116}			& 0.107	 /	\textbf{0.116}		&   0.111 	/	0.124	& \textbf{0.104} / 0.165		& 0.119  / \ding{55} \quad \quad \\
				indoor\_03 		&  0.296 / 0.873				& 0.071	/ \textbf{0.078}			& 0.072	 /	\textbf{0.078}			& \textbf{0.068} / 0.100	& 0.082 / 0.162		& 0.156 / 	\ding{55} \quad \quad \\
				outdoor\_01 	&  0.973 /	2.075				& 0.629  / 0.633		& 0.572	 /	0.631		&  0.835 /	0.731		& \textbf{0.533} / \textbf{0.604}	& 0.691 /	0.756 \\
				outdoor\_02 	&  \ding{55} / \ding{55}		& 0.097	/ \textbf{0.098}			& \textbf{0.096} / \textbf{0.098}	&  0.100  /	0.180		& 0.105  	 	/ 0.242		& 0.100   / 0.107\\
				outdoor\_03 	&  \ding{55} / \ding{55}		& 0.500 /  0.501		& 0.504 /	0.501		&  1.980 /	0.859		& \textbf{0.493} / \textbf{0.148} 	&  0.432 /	 0.327 \\
				outdoor\_05 	&  1.312 / \ding{55} \quad \quad& 0.821 /	0.823		& 0.819	 /	0.824		&  0.847  /	1.352		&  0.796  		/ 0.185		& \textbf{0.779} / \textbf{0.292}	 \\
				outdoor\_06 	&  3.434 / \ding{55} \quad	\quad& 0.163 /  0.170		& 0.164	 /	\textbf{0.169}		&  0.163  /	0.339		& 0.163  		/0.241		&  \textbf{0.130} / 0.233 	 \\
				\midrule
				Avg. 			& -  / -						 &0.341 	/ 0.346		&  0.333  /  \textbf{0.345}  & 0.586 / 0.526    & \textbf{0.325}/ 0.392 		& 0.344 / - \\
				\bottomrule[1.5pt]
			\end{tabular}
			Unit: [m]. The notation ``ev'' indicates results estimated using the IMU of the event camera, while ``uav'' indicates results derived from the IMU of the quadrotor.
		\end{threeparttable}
	}
\end{table}

Unlike the varying velocities in foregoing public datasets, the own-collected sequences typically have nearly constant velocities.  
Intuitively, it will result in low Signal-to-Noise Ratio (SNR) measurements for inertial sensors, as constant-speed motion normally induces weak stimulation for inertial sensors.
It will be helpful for further validating the sensitivity of different inertial schemes to high-quality and low-quality IMUs.
The ATE results are summarized in Table~\ref{The RMSE of Own Collected dataset}. 
Compared to using the quadrotor IMU, USLAM and $\text{GPP}^{\ast}$ achieved higher accuracy and success rate when using the camera IMU. Within the drone-IMU sequences (labeled as \emph{uav}), the performance of GPP is even worse than that of Preint. Conversely, the GPP can achieve the best accuracy using the camera IMU (labeled as \emph{ev}). On the other hand, Preint and ExtPreint exhibited less sensitivity to the quality of IMU measurements. These comparative results are consistent with the simulation analysis in Sec.~\ref{sec:analy_inertial_factors}. 
In fact, the WNOJ motion prior is more consistent with the constant-speed ground truth in these sequences, enabling the estimator to achieve a more accurate estimated trajectory (as shown in Fig~\ref{overall_trajectory_top_own_collected}). 
On the contrary, the GPP method relies solely on inertial measurements to construct an induced trajectory, making it more susceptible to the influence of low-stimulation, noisy inertial data. 
In summary, these low Signal-to-Noise Ratio (SNR) measurements from the IMU leads to a degenerate performance for GPP.
\begin{table}[!t]
	\centering
	\renewcommand{\arraystretch}{1.2}
	\caption{The Time Cost of Each Module [ms]}
	\label{time_cost_table}
	\resizebox{3.5in}{!}{
		\begin{threeparttable}
			\begin{tabular}{ l c c c c c c}
				\toprule[1.5pt]
				Module  		&  GPP		& GPIF 			& ExtPreint 	& DEVO & PL-EVIO & ESVIO \\
				\cmidrule(r){1-4} \cmidrule(r){5-7}
				\multirow{2}{*}{Event Front-End}		&  		\multicolumn{3}{c}{($3.78 \times 10^{-3}\ ms/e$,  $2.01 \times$ time lag, CPU)} & \multirow{2}{*}{24.52} & \multirow{2}{*}{8.68} & \multirow{2}{*}{35.69}\\
				& \multicolumn{3}{c}{($1.28 \times 10^{-4}\ ms/e$, $3.20 \times$ real-time, GPU)} & \\
				\cmidrule(r){1-4} \cmidrule(r){5-7}
				IMU Preintegration 	&  7.13			& 0.00				& 0.19            & 0.00 & - & -	\\
				\cmidrule(r){1-4} \cmidrule(r){5-7}
				Construct Residuals &   5.77     &   1.18           	&    0.88         & 1.32 &0.75& - \\
				\cmidrule(r){1-4} \cmidrule(r){5-7}
				Optimization	&  495.59			&  1012.48 			&   829.26          & 1.98 &37.66& 35.59\\
				\cmidrule(r){1-4} \cmidrule(r){5-7}
				Marginalization &  	4.05		&   11.95			&   8.27          & 5.94 &7.50& -\\
				\bottomrule[1.5pt]
			\end{tabular}
			The time cost results of PL-EVIO and ESVIO are taken from  \cite{Guan2023plevio} and \cite{chen2023esvio}, respectively. The results for DEVO are reproduced using their open-source code \cite{klenk2024deep}. Unreported time consumption is marked with ``$-$''.
		\end{threeparttable}
	}
\end{table}

\subsection{Runtime Analysis} 
To analyze time efficiency, we compare our AsynEIO with existing methods, including DEVO \cite{klenk2024deep}, PL-EVIO \cite{Guan2023plevio} and ESVIO \cite{chen2023esvio}, by counting the time-cost of each module (as summarized in Table~\ref{time_cost_table}). DEVO is a learning-based method where a patch network is leveraged to refine the target feature location according newest reprojection and a vectorized optimization back-end is specially designed for parallel inference. In DEVO, event streams are converted into voxel grids and reduced to a quarter of their original size. 
PL-EVIO and ESVIO have pipelines similar to the conventional VIO, which can enable real-time estimation without any hardware acceleration. The fully asynchronous front-end of AsynEIO can serially process each event in $3.78 \times 10^{-3}\ ms$. We observe an average lag of 2.01 times ($155.15\ s / 77.32\ s$) for a $346 \times 260$ resolution event camera on UZH-FPV dataset. To further improve the real-time performance, we also realize a fully parallel asynchronous event front-end where new tracker create and active tracker updating are assigned to different CUDA streams. Each feature trajectory is tracked by an individual thread block, within which multiple threads are utilized for parallel processing. In our parallel front-end, Both FA-Harris and HASTE are fully implemented using CUDA to ensure that the entire event front-end operates on the GPU. As shown in Table~\ref{time_cost_table}, our parallel front-end (labeled as GPU) achieves an asynchronous processing speed of $1.28 \times 10^{-4}\ ms$ per event. Although the total front-end processing time increases due to the bandwidth limitation between GPU and CPU memory, the overall front-end is still $3.20\ \times$ faster than the real-time requirement, and achieves a $6.42\ \times$ speedup compared to the CPU-based implementation. The IMU preintegration in GPP requires a relatively high time cost of $7.13\ ms$. Although GPIF has no cost in preintegration, it expends more computational effort in constructing residuals ($1.18\ ms$), execution optimization ($1012.48 ms$), and marginalization ($11.95\ ms$), as it generates new factors for each IMU measurement.
Conversely, GPP and ExtPreint only need to attach a single factor for each preintegration measurement. The high cost ($5.77\ ms$) for GPP primarily arises from querying the induced latent states trajectory during the asynchronous reprojection. 
The global GP trajectory of GPIF and ExtPreint has higher dimensions ($18+6\ dims$) than the local induced trajectory of GPP ($9+6\ dims$). Therefore, GPP achieves an efficient back-end for optimization and marginalization. 

\section{Discussion}
\label{sec:Discussion}

\subsection{Choice of Inertial Schemes}

Generally, these GP-trajectory-based methods (i.e. Preint, ExtPreint, GPIF, GPP*) would result in high-precision estimation, if the ground-truth trajectory is inherently smooth. In this case, these methods can even handle short-term measurement loss or tracking failure. For applications involving moving vehicles or drones, these solutions are advisable. 
At lower state frequency settings, GPIF generally demonstrates suboptimal performance, but it has a low-pass filtering effect to IMU measurements. When configured at higher state frequencies and stronger IMU stimulation, GPIF can achieve enhanced accuracy. Nevertheless, it sacrifices the time-efficiency. Thus, GPIF might be more suitable for scenarios where real-time performance is not a priority. For instance, offline computation of estimated trajectories from low-quality IMUs.
Conversely, the performance of GPP is primarily influenced by the quality of IMUs (frequency and noise) and data associations. Normally, GPP requires high SNR measurements and has minimal dependence on the motion pattern.
For example, GPP is particularly recommended for handheld walking sequences.  Furthermore, GPP stands out as the most computationally efficient method among them. Ultimately, ExtPreint can be considered a trade-off between accuracy and time efficiency.

\subsection{Limitations and Future Works}
Although the proposed system has shown competitive robustness and accuracy, it still is imperfect in several aspects. The main drawback would be the poor runtime performance of back-end. In the current GPU-based implementation, the asynchronous front-end has realized good real-time performance.
However, the factor graph in the back-end remains somewhat a large scale, especially for GPIF. The residual factors should be organized in a more tight structure rather than keeping abundant sole factors between the same state pairs. More strategic marginalization methods should be specifically designed for the GP-based back-end. Constrained by the performance of the event tracker itself, our front-end still exhibits significant tracking errors. More acceleration techniques and high-precision asynchronous trackers are anticipated for future work.

\section{Conclusion}
\label{sec:conclusion}

This article introduces an asynchronous event-inertial odometry called AsynEIO that leverages a unified GP regression framework, where several important inertial fusion schemes are realized and evaluated in the experiments. To the best of our knowledge, it is the first work that makes fair comparisons for asynchronous event-inertial fusion system. The expressing abilities of GP trajectories defined in different spaces and motion priors are assessed in detail, which can offer valuable suggestion for choosing the suitable one according to certain applications. The comparison analysis of various inertial factors demonstrates their properties. Experiments conducted on real event-inertial datasets with the state-of-the-art illustrate that AsynEIO has competitive accuracy and robustness.
Especially, our AsynEIO has a satisfied performance in high-speed maneuvering and low-illumination scenarios, which mainly benefited from the proposed asynchronous tracking and fusion mechanism. Improving the time-efficiency of AsynEIO is an area of future work.

\section*{Appendix}

\subsection{Perturbing the Pose on $SE(3)$ }
\begin{align}
	\boldsymbol{T} &= \bar{\boldsymbol{T}} \exp(\boldsymbol{\epsilon_{1}}^{\land}) = 
	\exp(\boldsymbol{\epsilon_{2}}^{\land}) \bar{\boldsymbol{T}} = \bar{\boldsymbol{T}} \exp(( \boldsymbol{\mathcal{T}}^{-1} \boldsymbol{\epsilon_{2}})^{\land}),
\end{align}
where $\boldsymbol{\mathcal{T}} = Ad(\exp(\boldsymbol{\xi}^{\land})) = \exp(\boldsymbol{\xi}^{\curlywedge}) $ is the adjoint of $SE(3)$.

\subsection{Perturbing WNOJ Prior Residuals}
\begin{align}
	\boldsymbol{\xi}_{k,k+1} &=  \log(\boldsymbol{T}_k^{-1}\boldsymbol{T}_{k+1})^\vee \nonumber\\
	&\approx \bar{\boldsymbol{\xi}}_{k,k+1} + \underbrace{ \boldsymbol{\mathcal{J}}_{k,k+1,op}^{-1} (\boldsymbol{\epsilon}_{k+1} - \boldsymbol{\mathcal{T}}_{k,k+1,op}^{-1} \boldsymbol{\epsilon}_{k}) }_{\triangleq \delta \boldsymbol{\xi}_{k,k+1}} \\
	\boldsymbol{\mathcal{J}}^{-1}_{k,k+1} &= \boldsymbol{\mathcal{J}}^{-1}(\boldsymbol{\xi}_{k,k+1}) \nonumber\\
	&\approx \boldsymbol{\mathcal{J}}_{k,k+1,op}^{-1} + \frac{1}{2} (\delta \boldsymbol{\xi}_{k,k+1})^{\curlywedge} \\
\end{align}

\begin{align}
	\boldsymbol{\mathcal{J}}_{k,k+1}^{-1} \boldsymbol{\varpi}_{k+1} &= \left(\boldsymbol{\mathcal{J}}_{k,k+1,op}^{-1} + \frac{1}{2} (\delta \boldsymbol{\xi}_{k,k+1})^{\curlywedge}\right) \boldsymbol{\varpi}_{k+1} \nonumber\\
	&\approx \boldsymbol{\mathcal{J}}_{k,k+1,op}^{-1} \bar{\boldsymbol{\varpi}}_{k+1} + \boldsymbol{\mathcal{J}}_{k,k+1,op}^{-1} \delta \boldsymbol{\varpi}_{k+1} \nonumber\\
	&+ \underbrace{\frac{1}{2} (\delta \boldsymbol{\xi}_{k,k+1})^{\curlywedge} \bar{\boldsymbol{\varpi}}_{k+1}}_{-\frac{1}{2} \bar{\boldsymbol{\varpi}}_{k+1}^{\curlywedge} \delta \boldsymbol{\xi}_{k,k+1} } \\
	\boldsymbol{\mathcal{J}}_{k,k+1}^{-1} \dot{\boldsymbol{\varpi}}_{k+1} &= \left(\boldsymbol{\mathcal{J}}_{k,k+1,op}^{-1} + \frac{1}{2} (\delta \boldsymbol{\xi}_{k,k+1})^{\curlywedge}\right) \dot{\boldsymbol{\varpi}}_{k+1} \nonumber\\
	&\approx \boldsymbol{\mathcal{J}}_{k,k+1,op}^{-1} \bar{\dot{\boldsymbol{\varpi}}}_{k+1} + \boldsymbol{\mathcal{J}}_{k,k+1,op}^{-1} \delta \dot{\boldsymbol{\varpi}}_{k+1} \nonumber\\
	&+ \frac{1}{2} (\delta \boldsymbol{\xi}_{k,k+1})^{\curlywedge} \bar{\dot{\boldsymbol{\varpi}}}_{k+1}
\end{align}

\begin{align}
	(\boldsymbol{\mathcal{J}}_{k,k+1}^{-1} \boldsymbol{\varpi}_{k+1} )^{\curlywedge} \boldsymbol{\varpi}_{k+1} = (\boldsymbol{\mathcal{J}}_{k,k+1,op}^{-1} \bar{\boldsymbol{\varpi}}_{k+1} )^{\curlywedge} \bar{\boldsymbol{\varpi}}_{k+1} \nonumber\\
	+ (\boldsymbol{\mathcal{J}}_{k,k+1,op}^{-1} \bar{\boldsymbol{\varpi}}_{k+1} )^{\curlywedge} \delta \boldsymbol{\varpi}_{k+1}  \nonumber\\
	- \bar{\boldsymbol{\varpi}}_{k+1}^{\curlywedge} \boldsymbol{\mathcal{J}}_{k,k+1,op}^{-1} \delta \boldsymbol{\varpi}_{k+1}   \nonumber\\
	+ \underbrace{\left(\frac{1}{2} (\delta \boldsymbol{\xi}_{k,k+1})^{\curlywedge} \bar{\boldsymbol{\varpi}}_{k+1}\right)^{\curlywedge} \bar{\boldsymbol{\varpi}}_{k+1}}_{\frac{1}{2} \bar{\boldsymbol{\varpi}}_{k+1}^{\curlywedge} \bar{\boldsymbol{\varpi}}_{k+1}^{\curlywedge} \delta \boldsymbol{\xi}_{k,k+1}}
\end{align}

\subsection{Jacobians of Interpolated State w.r.t Sampling States}
\label{sec:ja_int_wrt_kin}
The relationship between the linear local state and kinematic state at sampling timestamps is
\begin{align}
	\boldsymbol{\gamma}_k(t_k) &= \begin{bmatrix}
		\boldsymbol{0}\\
		\boldsymbol{\varpi}_{k} \\
		\dot{\boldsymbol{\varpi}}_{k}
	\end{bmatrix}, \\
	\boldsymbol{\gamma}_k(t_{k+1}) &= \begin{bmatrix}
		\boldsymbol{\xi}_{k,k+1} \\
		\boldsymbol{\mathcal{J}}^{-1}_{k,k+1}  \boldsymbol{\varpi}_{k+1} \\
		\boldsymbol{\mathcal{J}}^{-1}_{k,k+1}  \dot{\boldsymbol{\varpi}}_{k+1} + \frac{1}{2} (\boldsymbol{\mathcal{J}}^{-1}_{k,k+1} \boldsymbol{\varpi}_{k+1})^{\curlywedge} \boldsymbol{\varpi}_{k+1}
	\end{bmatrix}.
\end{align}

The interpolated local state can be expressed with sampling states as
\begin{align}
	\label{eq:interp_linear_var}
	\boldsymbol{\xi}_{k,\tau} &= \boldsymbol{\xi}_{k}(\tau) \nonumber\\
	&= \boldsymbol{\Lambda}_{12}(\tau) \boldsymbol{\varpi}_{k} +  \boldsymbol{\Lambda}_{13}(\tau) \dot{\boldsymbol{\varpi}}_{k} + \boldsymbol{\Psi}_{11}(\tau) \boldsymbol{\xi}_{k,k+1} \nonumber\\  
	&+ \boldsymbol{\Psi}_{12}(\tau) \boldsymbol{\mathcal{J}}^{-1}_{k,k+1}  \boldsymbol{\varpi}_{k+1} + \boldsymbol{\Psi}_{13}(\tau) \boldsymbol{\mathcal{J}}^{-1}_{k,k+1}  \dot{\boldsymbol{\varpi}}_{k+1} \nonumber\\
	&+ \frac{1}{2} \boldsymbol{\Psi}_{13}(\tau) (\boldsymbol{\mathcal{J}}^{-1}_{k,k+1} \boldsymbol{\varpi}_{k+1})^{\curlywedge} \boldsymbol{\varpi}_{k+1} \\
	\boldsymbol{T}(\tau) &= \boldsymbol{T}_{k} \exp((\boldsymbol{\xi}_{k,\tau})^{\land})
\end{align}
Notice that $\dot{\boldsymbol{\xi}}_{k,\tau}$ and $\ddot{\boldsymbol{\xi}}_{k,\tau}$ have the same forms as $\boldsymbol{\xi}_{k,\tau}$, but have different rows of matrix coefficients. With \eqref{eq:rel_local_velocity} and \eqref{eq:ddxi}, we can resolve $\boldsymbol{\varpi}(\tau)$ and $\dot{\boldsymbol{\varpi}}(\tau)$.

We first derive the perturbing Jacobians of $\boldsymbol{T}(\tau)$ w.r.t $\boldsymbol{T}_{k}$ and $\boldsymbol{\xi}_{k,\tau}$, respectively. 
Then, the perturbing Jacobians of $\boldsymbol{\xi}_{k,\tau}$ w.r.t $\boldsymbol{x}(k)$ and $\boldsymbol{x}(k+1)$ are given by
\begin{align}
	\log(\boldsymbol{T}(\tau))^{\vee} &= \log(\bar{\boldsymbol{T}}(\tau) \exp(\boldsymbol{\epsilon}_{\tau}^{\land}))^{\vee}
	\approx \log(\bar{\boldsymbol{T}}(\tau))^{\vee} + \boldsymbol{\mathcal{J}}^{-1}  \boldsymbol{\epsilon}_{\tau} \\
	\log(\boldsymbol{T}(\tau))^{\vee} &= \log(\boldsymbol{T}_{k} \exp((\boldsymbol{\xi}_{k,\tau})^{\land}))^{\vee} \nonumber\\
	&\approx \log(\bar{\boldsymbol{T}}(\tau))^{\vee} + \boldsymbol{\mathcal{J}}^{-1} \boldsymbol{\mathcal{J}}(\bar{\boldsymbol{\xi}}_{k,\tau}) \delta \boldsymbol{\xi}_{\tau} \\ 
	\log(\boldsymbol{T}(\tau))^{\vee} &=
	\log(\bar{\boldsymbol{T}}_{k} \exp(\boldsymbol{\epsilon}_{k}^{\land}) \exp(\boldsymbol{\xi}_{k,\tau} ^{\land}))^{\vee} \nonumber\\
	&\approx\log(\bar{\boldsymbol{T}}(\tau))^{\vee} + \boldsymbol{\mathcal{J}}^{-1} (\boldsymbol{\mathcal{T}}(\boldsymbol{\xi}_{k,\tau}))^{-1} \boldsymbol{\epsilon}_{k}
\end{align}

According the chain rule, we can define the Jacobians of interpolated state w.r.t $\boldsymbol{T}_{k}$ and $\boldsymbol{\xi}_{k,\tau}$ as
\begin{align}
	\frac{\partial \boldsymbol{T}(\tau)}{\partial \boldsymbol{\xi}_{k,\tau}} &= \frac{\partial \boldsymbol{T}_{k} \exp((\bar{\boldsymbol{\xi}}_{k,\tau} + \delta \boldsymbol{\xi}_{\tau})^{\land}) }{\partial \delta \boldsymbol{\xi}_{\tau}} =  \boldsymbol{\mathcal{J}}(\bar{\boldsymbol{\xi}}_{k,\tau}), \\
	\frac{\partial \boldsymbol{T}(\tau)}{\partial \boldsymbol{T}_{k}} &= \frac{\partial \bar{\boldsymbol{T}}_{k} \exp(\boldsymbol{\epsilon}_{k}^{\land}) \exp(\boldsymbol{\xi}_{k,\tau} ^{\land}) }{\partial \boldsymbol{\epsilon}_{k}} =  (\boldsymbol{\mathcal{T}}(\boldsymbol{\xi}_{k,\tau}))^{-1}.
\end{align}
Perturbing the right-hand side of \eqref{eq:rel_local_velocity} with $\delta \boldsymbol{\xi}$ and $\delta \dot{\boldsymbol{\xi}}$, we have
\begin{align}
	\boldsymbol{\varpi}(\tau) &= \boldsymbol{\mathcal{J}}(\boldsymbol{\xi}_{k,\tau})  \dot{\boldsymbol{\xi}}_{k,\tau} \\
	&\approx \boldsymbol{\mathcal{J}}_{k,\tau,op} \bar{\dot{\boldsymbol{\xi}}}_{k,\tau} + \frac{1}{2} \bar{\dot{\boldsymbol{\xi}}}_{k,\tau}^{\curlywedge} \delta \boldsymbol{\xi} + \boldsymbol{\mathcal{J}}_{k,\tau,op} \delta \dot{\boldsymbol{\xi}}.
\end{align}
On this basis, the Jacobians of $\boldsymbol{\varpi}(\tau)$ w.r.t the local variables are given by
\begin{align}
	\frac{\partial \boldsymbol{\varpi}(\tau)}{\partial \boldsymbol{\xi}_{k,\tau}} = \frac{1}{2} \bar{\dot{\boldsymbol{\xi}}}_{k,\tau}^{\curlywedge}, \quad \frac{\partial \boldsymbol{\varpi}(\tau)}{\partial \dot{\boldsymbol{\xi}}_{k,\tau}} = \boldsymbol{\mathcal{J}}_{k,\tau,op}, \quad
	\frac{\partial \boldsymbol{\varpi}(\tau)}{\partial \ddot{\boldsymbol{\xi}}_{k,\tau}} &= \boldsymbol{0}
\end{align}
Calculating the derivative on both sides of \eqref{eq:rel_local_velocity} yields
\begin{align}
	\dot{\boldsymbol{\varpi}}(\tau) &= \boldsymbol{\mathcal{J}}_{k,\tau} \ddot{\boldsymbol{\xi}}_{k,\tau} + \frac{d}{dt}\boldsymbol{\mathcal{J}}_{k,\tau} \dot{\boldsymbol{\xi}}_{k,\tau}
\end{align}
Therefore, we have the Jacobians as
\begin{align}
	\frac{\partial \dot{\boldsymbol{\varpi}}(\tau)}{\partial \boldsymbol{\xi}_{k,\tau}} = \frac{1}{2} \bar{\ddot{\boldsymbol{\xi}}}_{k,\tau}^{\curlywedge}, \quad
	\frac{\partial \dot{\boldsymbol{\varpi}}(\tau)}{\partial \dot{\boldsymbol{\xi}}_{k,\tau}} = \boldsymbol{0}, \quad
	\frac{\partial \dot{\boldsymbol{\varpi}}(\tau)}{\partial \ddot{\boldsymbol{\xi}}_{k,\tau}} = \boldsymbol{\mathcal{J}}_{k,\tau,op}
\end{align}
Perturbing the right-hand side of \eqref{eq:interp_linear_var} with $\delta \boldsymbol{\varpi}_{k}$, $\delta \boldsymbol{\varpi}_{k+1}$ and $\delta \boldsymbol{\xi}_{k,k+1}$,  we have
\begin{align}
	\boldsymbol{\xi}_{k,\tau} &= \boldsymbol{\xi}_{k,\tau,op} + \boldsymbol{\Lambda}_{12} \delta \boldsymbol{\varpi}_{k} + \boldsymbol{\Lambda}_{13} \delta \dot{\boldsymbol{\varpi}}_{k} + \boldsymbol{\Psi}_{11}(\tau) \delta \boldsymbol{\xi}_{k,k+1} \nonumber\\
	&+ \boldsymbol{\Psi}_{12}(\tau)  \boldsymbol{\mathcal{J}}_{k,k+1,op}^{-1} \delta \boldsymbol{\varpi}_{k+1} -\frac{1}{2} \boldsymbol{\Psi}_{12}(\tau)  \bar{\boldsymbol{\varpi}}_{k+1}^{\curlywedge} \delta \boldsymbol{\xi}_{k,k+1} \nonumber\\
	&+ \boldsymbol{\Psi}_{13}(\tau)  \boldsymbol{\mathcal{J}}_{k,k+1,op}^{-1} \delta \dot{\boldsymbol{\varpi}}_{k+1} -\frac{1}{2} \boldsymbol{\Psi}_{13}(\tau)  \bar{\dot{\boldsymbol{\varpi}}}_{k+1}^{\curlywedge} \delta \boldsymbol{\xi}_{k,k+1} \nonumber\\
	&+ \frac{1}{2} \boldsymbol{\Psi}_{13}(\tau) (\boldsymbol{\mathcal{J}}_{k,k+1,op}^{-1} \bar{\boldsymbol{\varpi}}_{k+1} )^{\curlywedge} \delta \boldsymbol{\varpi}_{k+1} \nonumber\\
	&- \frac{1}{2} \boldsymbol{\Psi}_{13}(\tau) \bar{\boldsymbol{\varpi}}_{k+1}^{\curlywedge} \boldsymbol{\mathcal{J}}_{k,k+1,op}^{-1} \delta \boldsymbol{\varpi}_{k+1} \nonumber\\
	&+ \frac{1}{4} \boldsymbol{\Psi}_{13}(\tau)  \bar{\boldsymbol{\varpi}}_{k+1}^{\curlywedge} \bar{\boldsymbol{\varpi}}_{k+1}^{\curlywedge} \delta \boldsymbol{\xi}_{k,k+1}
\end{align}
Therefore, we have Jacobians as follows:
\begin{align}
	\frac{\partial \boldsymbol{\xi}_{k,\tau}}{\partial \boldsymbol{\epsilon}_{k+1}} &= \boldsymbol{\Psi}_{11}(\tau) \boldsymbol{\mathcal{J}}_{k,k+1,op}^{-1}  \nonumber \\
	&-\frac{1}{2} \boldsymbol{\Psi}_{12}(\tau)  \bar{\boldsymbol{\varpi}}_{k+1}^{\curlywedge} \boldsymbol{\mathcal{J}}_{k,k+1,op}^{-1}  \nonumber\\
	&- \frac{1}{2} \boldsymbol{\Psi}_{13}(\tau)  \bar{\dot{\boldsymbol{\varpi}}}_{k+1}^{\curlywedge} \boldsymbol{\mathcal{J}}_{k,k+1,op}^{-1}  \nonumber \\
	&+\frac{1}{4} \boldsymbol{\Psi}_{13}(\tau) \bar{\boldsymbol{\varpi}}_{k+1}^{\curlywedge} \bar{\boldsymbol{\varpi}}_{k+1}^{\curlywedge} \boldsymbol{\mathcal{J}}_{k,k+1,op}^{-1}  \\
	\frac{\partial \boldsymbol{\xi}_{k,\tau}}{\partial \boldsymbol{\epsilon}_{k}} &= -\frac{\partial \boldsymbol{\xi}_{k,\tau}}{\partial \boldsymbol{\epsilon}_{k+1}} \boldsymbol{\mathcal{T}}_{k,k+1,op}^{-1}, \quad \frac{\partial \boldsymbol{\xi}_{k,\tau}}{\partial \delta \boldsymbol{\varpi}_{k}} = \boldsymbol{\Lambda}_{12} \\
	\frac{\partial \boldsymbol{\xi}_{k,\tau}}{\partial \delta \boldsymbol{\varpi}_{k+1}} &= \boldsymbol{\Psi}_{12}(\tau)  \boldsymbol{\mathcal{J}}_{k,k+1,op}^{-1} \nonumber \\
	&+\frac{1}{2} \boldsymbol{\Psi}_{13}(\tau) (\boldsymbol{\mathcal{J}}_{k,k+1,op}^{-1} \bar{\boldsymbol{\varpi}}_{k+1} )^{\curlywedge} \nonumber \\
	&- \frac{1}{2} \boldsymbol{\Psi}_{13}(\tau) \bar{\boldsymbol{\varpi}}_{k+1}^{\curlywedge} \boldsymbol{\mathcal{J}}_{k,k+1,op}^{-1} \\
	\frac{\partial \boldsymbol{\xi}_{k,\tau}}{\partial \delta \dot{\boldsymbol{\varpi}}_{k}} &= \boldsymbol{\Lambda}_{13} \quad
	\frac{\partial \boldsymbol{\xi}_{k,\tau}}{\partial \delta \dot{\boldsymbol{\varpi}}_{k+1}} = \boldsymbol{\Psi}_{13}(\tau)  \boldsymbol{\mathcal{J}}_{k,k+1,op}^{-1}
\end{align}

\bibliographystyle{IEEEtran}
\bibliography{IEEEabrv,mybibfile}


\end{document}